\documentclass[10pt,twocolumn,letterpaper]{article}

\usepackage{cvpr}
\usepackage{times}
\usepackage{epsfig}
\usepackage{graphicx}
\usepackage{amsmath}
\usepackage{amssymb}
\usepackage{svg}
\usepackage{color}
\usepackage{subfig}
\usepackage{authblk}
\usepackage{xcolor}
\usepackage[section]{placeins}

\usepackage[pagebackref=true,breaklinks=true,letterpaper=true,colorlinks,bookmarks=false]{hyperref}

\cvprfinalcopy 

\ifcvprfinal\fi

\begin{document}

\title{Improved Fusion of Visual and Language Representations \\
by Dense Symmetric Co-Attention for Visual Question Answering}

\author{Duy-Kien Nguyen$^{1}$~ and ~~Takayuki Okatani$^{1,2}$\\
\vspace*{-3mm}
$^1$Tohoku University ~~~~ $^2$RIKEN Center for AIP\\
{\tt\small \{kien, okatani\}@vision.is.tohoku.ac.jp}}

\maketitle
\thispagestyle{empty}

\begin{abstract}
A key solution to visual question answering (VQA) exists in how to fuse visual and language features extracted from an input image and question. We show that 
an attention mechanism that enables dense, bi-directional interactions between the two modalities 
contributes to boost accuracy of prediction of answers. Specifically, we present a simple architecture that is fully symmetric between visual and language representations, in which each question word attends on image regions and each image region attends on question words. It can be stacked to form a hierarchy for multi-step interactions between an image-question pair.
We show through experiments that the proposed architecture achieves a new state-of-the-art on VQA and VQA 2.0 despite its small size. We also present qualitative evaluation, demonstrating how the proposed attention mechanism can generate reasonable attention maps on images and questions, which leads to the correct answer prediction. 
\end{abstract}

\section{Introduction}
There has been a significant progress in the study of visual question answering (VQA) over a short period of time since its introduction, showing rapid boost of performance for common benchmark datasets. This progress has been mainly brought about by two lines of research, 
the development of better attention mechanisms and the improvement in fusion of features extracted from an input image and question.

Since introduced by Bahdanau et al. \cite{Bahdanau}, attention has been playing an important role in solutions of various problems of artificial intelligence ranging from tasks using single modality (e.g., language, speech, and vision) to multimodal tasks. For VQA, attention on image regions generated from the input question was first introduced \cite{DBLP:conf/cvpr/YangHGDS16} and then several extensions have been proposed \cite{DBLP:conf/nips/LuYBP16, multi-level-attention-networks, chen2017sva}.
Meanwhile, researchers have proposed  several methods for feature fusion  
\cite{DBLP:conf/emnlp/FukuiPYRDR16,Kim2016c,Yu2017},
where the aim is to obtain better fused representation of image and question pairs. These studies updated the state-of-the-art  for common benchmark datasets at the time of each publication.

We observe that these two lines of research have been independently conducted so far. This is particularly the case with the studies of feature fusion methods, where attention is considered to be optional, even though the best performance is  achieved with it. However, we think that they are rather {\em two different approaches towards the same goal}. In particular, we argue that a better attention mechanism leads to a better fused representation of image-question pairs. 

Motivated by this, we propose a novel co-attention mechanism for improved fusion of visual and language representations. Given representations of an image and a question, it  first generates an attention map on image regions for each question word and an attention map on question words for each image region.  
It then performs computation of attended features, concatenation of multimodal representations, and their transformation by a single layer network with ReLU and a residual connection. 
These computations are encapsulated into a composite network that we call {\em dense co-attention} layer, since it considers every interaction between any image region and any question word. The layer has fully symmetric architecture between the two modalities, and can be stacked to form a hierarchy that enables multi-step interactions between the image-question pair. 

Starting from initial representations of an input image and question, each dense co-attention layer in the layer stack updates the representations, which are inputted to the next layer. Its final output are then fed to a layer for answer prediction. We use additional attention mechanisms in the initial feature extraction as well as the answer prediction layer. We call the entire network including all these components the {\em dense coattention network} (DCN). We show the effectiveness of DCNs by several experimental results; they achieve the new state-of-the-art for VQA 1.0 and 2.0 datasets.

\section{Related Work}

In this section, we briefly review previous studies of VQA with a special focus on the developments of attention mechanisms and fusion methods.

\subsection{Attention Mechanisms}

Attention has proved its effectiveness on many tasks and VQA is no exception. A number of methods have been developed so far, in which question-guided attention on image regions is commonly used. They are categorized into two classes according to the type of employed image features. One is the class of methods that use visual features from some region proposals, which are generated by Edge Boxes \cite{shih2016wtl, DBLP:journals/corr/IlievskiYF16} or Region Proposal Network \cite{DBLP:journals/corr/abs-1708-02711}. 
The other is the class of methods that use convolutional features (i.e., activations of convolutional layers) \cite{chen2017sva, DBLP:conf/emnlp/FukuiPYRDR16, DBLP:journals/corr/KazemiE17, kim2016b, Kim2016c, DBLP:conf/nips/LuYBP16, DBLP:journals/corr/NamHK16, DBLP:journals/corr/NohH16, Xu2016, DBLP:conf/cvpr/YangHGDS16, Yu2017}.

There are several approaches to creation and use of attention maps. Yang \etal~\cite{DBLP:conf/cvpr/YangHGDS16} developed stacked attention network that produces multiple attention maps on the image in a sequential manner, aiming at performing multiple steps of reasoning. Kim \etal~\cite{kim2016b} extended this idea by incorporating it into a residual architecture to produce better attention information. Chen \etal~\cite{chen2017sva} proposed a structured attention model 
that can encode cross-region relation, aiming at properly answering questions that involve complex inter-region relations.

Earlier studies mainly considered question-guided attention on image regions. In later studies, the opposite orientation of attention, i.e., image-guided attention on question words, is considered additionally. Lu \etal~\cite{DBLP:conf/nips/LuYBP16} introduced the co-attention mechanism that generates and uses attention on image regions and on question words. To reduce the gap of image and question features, Yu \etal~ \cite{multi-level-attention-networks} utilized attention to extract not only spatial information but also language concept of the image.
Yu \etal~\cite{Yu2017} combined the mechanism with a novel multi-modal feature fusion of image and question. 

We point out that the existing attention mechanisms only consider a limited amount of possible interactions between image regions and question words. Some consider only attention on image regions from a {\em whole} question. Co-attention additionally considers attention on question words but it is created from a {\em whole} image. We argue that this can be a significant limitation of the existing approaches. The proposed mechanism can deal with every interaction between any image region and any question word, which possibly enables to model unknown complex image-question relations  that are necessary for correctly answering questions.

\subsection{Multimodal Feature Fusion} 

The common framework of existing methods is that visual and language features are independently extracted from the image and question at the initial step, and they are fused at a later step to compute the final prediction.  
In early studies, researchers employed simple fusion methods such as the concatenation, summation, and element-wise product of the visual and language features, which are fed to fully connected layers to predict answers. 

It was then shown by Fukui \etal~\cite{DBLP:conf/emnlp/FukuiPYRDR16} that a more complicated fusion method does improve prediction accuracy; they introduced the bilinear (pooling) method that uses an outer product of two vectors of visual and language features for their fusion. As the outer product gives a very high-dimensional feature, they adopt the idea of Gao \etal~\cite{DBLP:journals/corr/GaoBZD15} to compress the fused feature and name it the Multimodal Compact Bilinear (MCB) pooling method. 
However, the compacted feature of the MCB method still tends to be high-dimensional to guarantee robust performance, Kim \etal~\cite{Kim2016c} proposed low-rank bilinear pooling using Hadamard product of two feature vectors, which is called the Multimodal Low-rank Bilinear (MLB) pooling. Pointing out that MLB suffers from slow convergence rate, Yu \etal~\cite{Yu2017} proposed the Multi-modal Factorized Bilinear (MFB) pooling, which computes a fused feature with a matrix factorization trick to reduce the number of parameters and improve convergence rate.

The attention mechanisms can also be considered feature fusion methods, regardless of whether it is explicitly mentioned, since they are designed to obtain a better representation of image-question pairs based on their interactions. This is particularly the case with co-attention mechanisms in which the two features are treated symmetrically. 
Our dense co-attention network is based on this observation. It fuses the two features by multiple applications of the attention mechanism that can use more fine-grained interactions between them. 

\section{Dense Co-Attention Network (DCN)}

\begin{figure}[t]
\centering
\includegraphics[width=80mm]{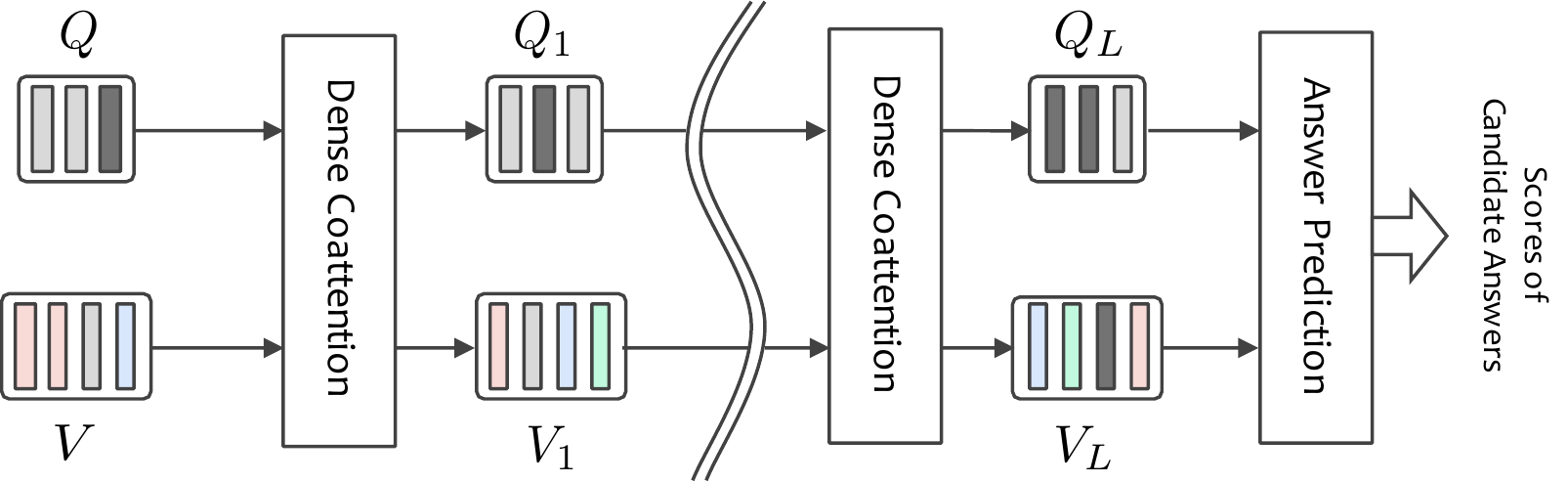}
\caption{The global structure of the dense co-attention network (DCN).}
\label{fig:global}
\end{figure}
In this section, we describe the architecture of DCNs; see Fig.\ref{fig:global} for its overview. It consists of a stack of {\em dense co-attention layers} that fuses language and visual features repeatedly, on top of which an {\em answer prediction layer} that predict answers in a multi-label classification setting \cite{DBLP:journals/corr/abs-1708-02711}. We first explain the initial feature extraction from the input question and image (Sec.\ref{sec:feat}) and then describe the dense co-attention layer (Sec.\ref{sec:coatt}) and the answer prediction layer (Sec.\ref{sec:anspred}).

\subsection{Feature Extraction}
\label{sec:feat}

We employ pretrained networks that are commonly used in previous studies \cite{kim2016b, Yosinski:2014:TFD:2969033.2969197, Kim2016c, chen2017sva} for encoding or extracting features from images, questions, and answers, such as pretrained ResNet \cite{DBLP:conf/cvpr/HeZRS16} with some differences from earlier studies.  

\subsubsection{Question and Answer Representation} 
\label{sec:langfeat}

We use bi-directional LSTM for encoding questions and answers. 
Specifically, 
a question consisting of $N$ words is first converted into a sequence $\{e^Q_1, ..., e^Q_N\}$ of GloVe vectors \cite{pennington2014glove}, which are then inputted into a 
one-layer bi-directional LSTM (Bi-LSTM) with a residual connection as
\begin{equation}
\label{eqn:qn1}
\overrightarrow{q_n} = \text{Bi-LSTM}(\overrightarrow{q_{n-1}}, e^Q_n),
\end{equation}
\begin{equation}
\label{eqn:qn2}
\overleftarrow{q_n} = \text{Bi-LSTM}(\overleftarrow{q_{n+1}}, e^Q_n).
\end{equation}
We then create a matrix $Q = [q_1, ..., q_N] \in \mathbb{R}^{d \times N}$ where $q_n = [\overrightarrow{q_n}^\top, \overleftarrow{q_n}^\top]^\top$ ($n=1,\ldots,N$).
We will also use $s_Q = [\overrightarrow{q_N}^\top, \overleftarrow{q_1}^\top]^\top$, concatenation of the last hidden states in the two paths, for obtaining representation of an input image (Sec.\ref{sec:imfeat}). We randomly initialized the Bi-LSTM. It is worth noting that we initially used a pretrained two-layer Bi-LSTM that yields Context Vectors (CoVe) in \cite{McCann2017LearnedIT}, which we found does not contribute to performance.

We follow a similar procedure to encode answers. An answer of $M$ words is converted into $\{e^A_1, ..., e^A_M\}$ and then inputted to the same Bi-LSTM, yielding the hidden states
$\overrightarrow{a_m}$ and $\overleftarrow{a_m}$ $(m=1,\ldots,M)$. 
We will use $s_A = [\overrightarrow{a_M}^\top, \overleftarrow{a_1}^\top]^\top$ for answer representation.

\subsubsection{Image Representation}
\label{sec:imfeat}

As in many previous studies, we use a pretrained CNN (i.e., a ResNet \cite{DBLP:conf/cvpr/HeZRS16} with 152 layers pretrained on 
ImageNet) to extract visual features of multiple image regions, but 
our extraction method is slightly different. We extract features from four conv. layers and then use a question-guided attention on these layers to fuse their features. We do this to exploit the maximum potential of the subsequent dense co-attention layers. We conjecture that features at different levels in the hierarchy of visual representation \cite{Zeiler2014, Yosinski_understandingneural} will be necessary to correctly answer a wide range of questions. 

To be specific, we extract outputs from the four conv. layers (after ReLU)
before the last four pooling layers. These are tensors of different sizes (i.e., $256\times112\times112$, $512\times56\times56, 1024\times28\times28$, and $2048\times14\times14$) and are converted into tensors of the same size  $(d\times 14\times 14)$  by applying max pooling with a different pooling size and one-by-one convolution to each. We also apply $l_2$ normalization on the depth dimension of each tensor as in \cite{VQA}. We reshape the normalized tensors into four $d\times T$ matrices, where $T=14\times 14$. 

Next, attention on the four layers is created from $s_Q$, the representation of the whole question defined above. We use a two-layer neural network having 724 hidden units with ReLU non-linearity to project $s_Q$ to the scores of the four layers as
\begin{equation}
[s_1, s_2, s_3, s_4] = \text{MLP}(s_Q),
\end{equation}
which are then normalized by softmax to obtain four attention weights $\alpha_1,\ldots,\alpha_4$. The weighted sum of the above four matrices is computed, yielding
a $d\times T$ matrix $V = [v_1, ..., v_T]$,
which is our representation of the input image. It stores the image feature at the $t$-th image region in its $t$-th column vector of size $d$.

\subsection{Dense Co-Attention Layer}
\label{sec:coatt}
\subsubsection{Overview of the Architecture}
\begin{figure}[t]
\centering
\includegraphics[width=83mm]{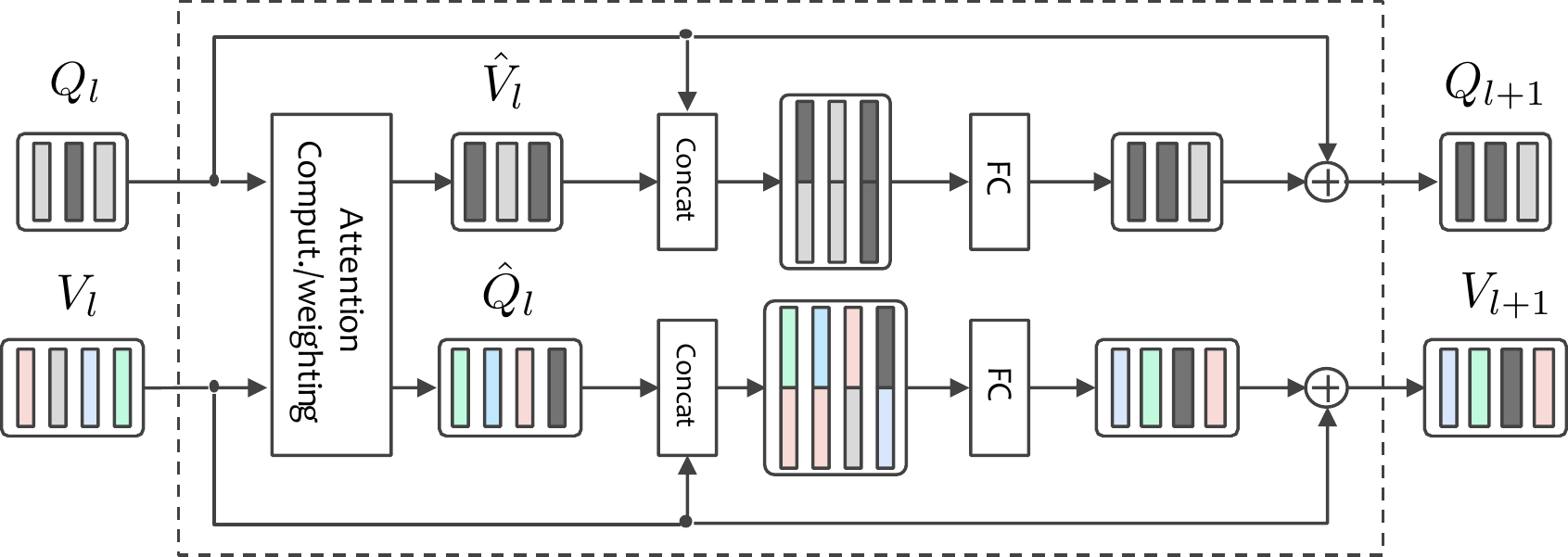}
\caption{The internal structure of a single dense co-attention layer of layer index  $l+1$.}
\label{fig:denselayer}
\end{figure}
We now describe the proposed dense co-attention layer; see Fig.\ref{fig:denselayer}. It takes the question and image representations $Q$ and $V$ as inputs and then outputs their updated versions. We denote the inputs to the $(l+1)$-st layer by $Q_l = [q_{l1}, ..., q_{lN}] \in \mathbb{R}^{d \times N}$ and $V_l = [v_{l1}, ..., v_{lT}] \in \mathbb{R}^{d \times T}$.
For the first layer inputs, we set $Q_0 = Q = [q_1, ..., q_N]$ and $V_0=V=[v_1, ..., v_T]$.

The proposed architecture has the following  properties. First, it is a co-attention mechanism \cite{DBLP:conf/nips/LuYBP16}.
Second, the co-attention is {\em dense} in the sense that it considers every interaction between any word and any region. To be specific, our mechanism creates one attention map on regions per each word and creates one attention map on words per each region (see Fig.\ref{fig:denseatt}). Third, it can be stacked as shown in Fig.\ref{fig:global}.  

\subsubsection{Dense Co-attention Mechanism}

\begin{figure}[t]
\centering
\includegraphics[width=65mm]{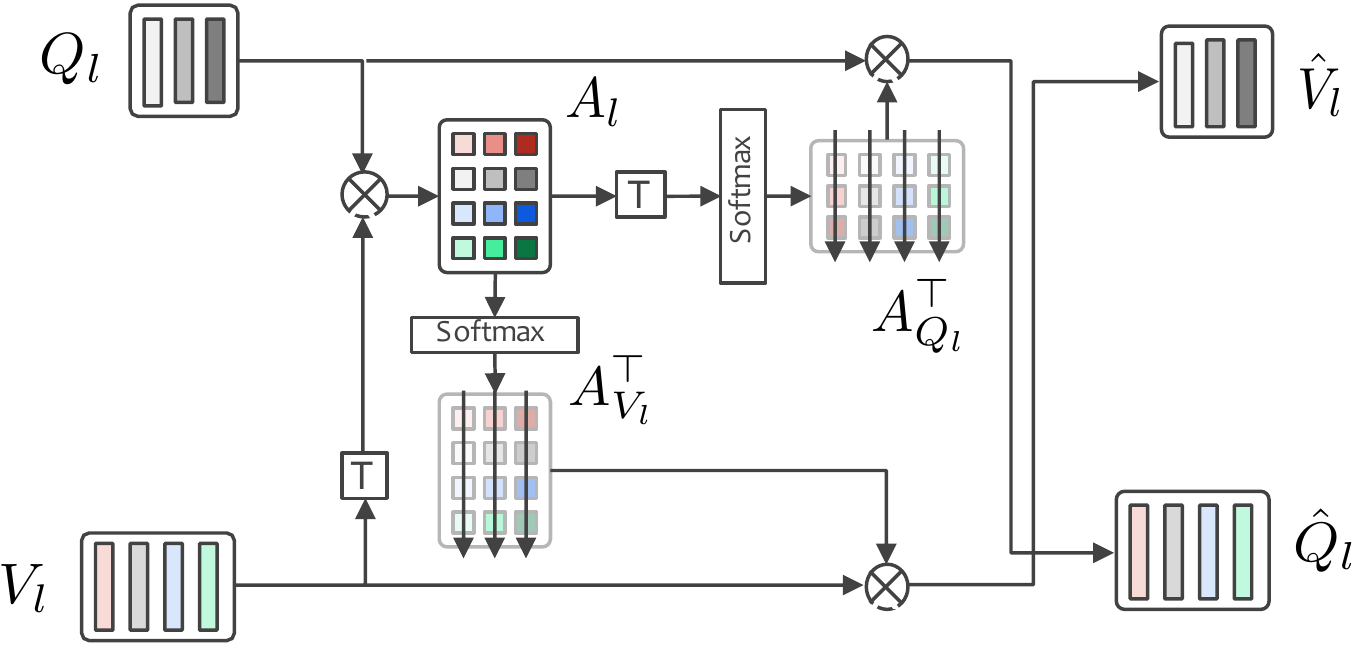}
\caption{Computation of dense co-attention maps and attended representations of the image and question. }
\label{fig:denseatt}
\end{figure}

\paragraph{Basic method for attention creation} For the sake of explanation,
we first explain the basic method for creation of attention maps, which we will  extend later.
Given $Q_l$ and $V_l$, 
two attention maps are created as shown in Fig.\ref{fig:denseatt}. Their computation  starts with the affinity matrix
\begin{equation}
A_l = V_l^{\top} W_l Q_l,
\end{equation}
where $W_l$ is a learnable weight matrix. 
We normalize $A_l$ in row-wise to derive attention maps on question words conditioned by each image region as
\begin{equation}
A_{Q_l} = \text{softmax}(A_l), 
\end{equation}
and also normalize $A_l$ in column-wise to derive attention maps on image regions conditioned by each question word as
\begin{equation}
A_{V_l} = \text{softmax}(A_l^{\top}).
\end{equation}
Note that each row of $A_{Q_l}$ and $A_{V_l}$ contains a single attention map.

\paragraph{Nowhere-to-attend and memory}
It often occurs at the creation and application of each attention map that there is no particular region or word that the model should attend. To deal with such cases, we add $K$ elements to $N$ question words as well as to $T$ image regions, as in \cite{XiongMS16}. In \cite{XiongMS16}, the authors only use $K = 1$, but we found it effective to use $K > 1$, which is expected to additionally serve as 
a memory for storing useful information \cite{DBLP:journals/corr/GravesWD14}.  More specifically, incorporating two matrices $M_{Q_l} \equiv [q_{l\oslash_1}, ..., q_{l\oslash_K}] \in \mathbb{R}^{d \times K}$ and $M_{V_l} \equiv [v_{l\oslash_1}, ..., v_{l\oslash_K}] \in \mathbb{R}^{d \times K}$, which are learnable parameters, we augment the matrix $Q_l$ and $V_l$  in the row direction as $\tilde{Q}_l = [q_{l1}, ..., q_{lN}, q_{l\oslash_1}, ..., q_{l\oslash_K}]
\in \mathbb{R}^{d \times (N+K)}$ and $\tilde{V}_l = [v_{l1}, ..., v_{lT}, v_{l\oslash_1}, ..., 
v_{l\oslash_K}] \in \mathbb{R}^{d \times (T+K)}$.
This augmentation of $Q_l$ and $V_l$ provides $A_l$ of size $(T+K)\times (N+K)$; $A_{Q_l}$ and $A_{V_l}$ are of size $(T+K)\times (N+K)$ and $(N+K)\times (T+K)$, respectively.


\paragraph{Parallel attention}
In several studies \cite{DBLP:journals/corr/KazemiE17, DBLP:journals/corr/VaswaniSPUJGKP17}, multiple attention maps are created and applied to target features in a parallel manner, which provides multiple attended features, and then they are {\em fused by concatenation}. In \cite{DBLP:journals/corr/VaswaniSPUJGKP17}, features are first linearly projected to multiple lower-dimensional spaces, for each of which the above attention function is performed. We adopt a similar approach that uses multiple attention maps here, but we use {\em average} instead of concatenation for fusion of the multiple attended features, because we found it works better in our case. 

To be specific, we linearly project the $d$-dimensional features (stored in the columns) of $\tilde{V}_l$ and $\tilde{Q}_l$ to multiple lower dimensional spaces. Let $h$ be the number of lower dimensional spaces and $d_h(\equiv d/h)$ be their dimension. We denote the linear projections by  $W_{\tilde{V}_l}^{(i)}\in \mathbb{R}^{d_h\times d}$ and $W_{\tilde{Q}_l}^{(i)}\in \mathbb{R}^{d_h\times d}$ $(i=1,\ldots,h)$. Then the affinity matrix between the projected features in the $i$-th space is given as
\begin{equation}
A_l^{(i)} = (W_{\tilde{V}_l}^{(i)}\tilde{V}_l)^\top(W_{\tilde{Q}_l}^{(i)}\tilde{Q}_l).
\end{equation} 
Attention maps are created from each affinity matrix by column-wise and row-wise normalization as
\begin{align}
A^{(i)}_{Q_l} &= \text{softmax}\left(\frac{A^{(i)}_l}{\sqrt{d_h}}\right),\\
A^{(i)}_{V_l} &= \text{softmax}\left(\frac{A^{(i)\top}_l}{\sqrt{d_h}}\right).
\end{align}
As we employ multiplicative (or dot-product) attention as explained below, average fusion of multiple attended features is equivalent to averaging our attention maps as
\begin{align}
A_{Q_l} &= \frac{1}{h}\sum_{i=1}^h A^{(i)}_{Q_l},\\
A_{V_l} &= \frac{1}{h}\sum_{i=1}^h A^{(i)}_{V_l}.
\end{align}

\paragraph{Attended feature representations}
We employ multiplicative attention 
to derive attended feature representations $\hat{Q}_l$ and $\hat{V}_l$ of the question and image, as shown in Fig.\ref{fig:denseatt}. As $A_{Q_l}$ and $A_{V_l}$ store attention maps in their rows and their last $K$ rows correspond to ``nowhere-to-attend" or memory, 
we discard them when applying them to $\tilde{Q}_l$ and $\tilde{V}_l$ as
\begin{equation}
\hat{Q}_l = \tilde{Q}_l \, A_{Q_l}\mathtt{ [1\!:\!T,:\,]}^\top, 
\end{equation}
and
\begin{equation}
\hat{V}_l = \tilde{V}_l \, A_{V_l}\mathtt{[1\!:\!N,:\,]}^\top, 
\end{equation}
respectively\footnote{The notation $\mathtt{(1\!:\!T,:\,)}$ indicates the submatrix in the first $T$ rows, as in Python.}. Note that $\hat{Q}_l$ is the same size as $V_l$ (i.e. $d\times T$) and $\hat{V}_l$ is the same size as $Q_l$ (i.e. $d\times N$).

\subsubsection{Fusing Image and Question Representations}

After computing the attended feature representations $\hat{Q}_l$ and $\hat{V}_l$, we fuse the image and question representations, as shown in the right half of Fig.\ref{fig:denselayer}. The matrix $\hat{V}_l$ stores  in its $n$-th column the attended representation of the entire image conditioned on the $n$-th question word. Then, the $n$-th column vector $\hat{v}_{ln}$ is fused with the representation $q_{ln}$ of $n$-th question word by concatenation to form $2d$-vector $[q_{ln}^\top,\hat{v}_{ln}^\top]^\top$.
This concatenated vector is projected back to a $d$-dimensional space by a single layer network followed by the ReLU activation and residual connection as
\begin{equation}
q_{(l+1)n} = \text{ReLU}\left(W_{Q_l}\begin{bmatrix}q_{ln}\\ \hat{v}_{ln}\end{bmatrix} + b_{Q_l}\right) + q_{ln},
\end{equation}
where $W_{Q_l}\in \mathbb{R}^{d \times 2d}$ and $b_{Q_l} \in\mathbb{R}^d$ are learnable weights and biases. An identical network (with the same weights and biases) is applied to each question word ($n=1,\ldots,N$) independently, yielding $Q_{l+1}=[q_{(l+1)1},\ldots,q_{(l+1)N}] \in \mathbb{R}^{d\times N}$.

Similarly, the representation $v_{lt}$ of $t$-th image region is concatenated with the representation $\hat{q}_{lt}$ of the whole question words conditioned on the $t$-th image region, and then projected back to a $d$-dimensional space as
\begin{equation}
v_{(l+1)t} = \text{ReLU}\left(W_{V_l}\begin{bmatrix}v_{lt}\\ \hat{q}_{lt}\end{bmatrix} + b_{V_l}\right) + v_{lt},
\end{equation}
where $W_{V_l} \in \mathbb{R}^{d\times 2d}$ and $b_{V_l}\in\mathbb{R}^d$ are weights and biases. The application of an identical network to each image region ($t=1,\ldots,T$) yields 
$V_{l+1}=[v_{(l+1)1},\ldots,v_{(l+1)T}] \in \mathbb{R}^{d\times T}$.

It should be noted that the above two fully-connected networks have different parameters (i.e., $W_{Q_l}$, $W_{V_l}$ etc.) for each layer $l$.

\subsection{Answer Prediction}
\label{sec:anspred}

Given the final outputs $Q_L$ and $V_L$ of the last dense co-attention layer, we predict answers. As they contain the representation of $N$ question words and $T$ image regions, we first perform self-attention function on each of them to obtain aggregated representations of the whole question and image. This is done for $Q_L$ as follows: 
i) compute  `scores' $s_{q_{L1}},\ldots,s_{q_{LN}}$ of $q_{L1},\ldots,q_{LN}$ 
by applying an identical two-layer MLP with ReLU nonlinearity in its hidden layer; ii) then apply softmax to them to derive attention weights $\alpha^Q_1,\ldots,\alpha^Q_N$; and iii) compute an aggregated representation by 
$s_{Q_L}=\sum_{n=1}^N \alpha^Q_n q_{Ln}$. Following the same procedure with an MLP with different weights, we derive attention weights $\alpha^V_1,\ldots,\alpha^V_T$ and then compute an aggregated representation $s_{V_L}$ from $v_{L1},\ldots,v_{LT}$. 

Using $s_{Q_L}$ and $s_{V_L}$ thus computed, we predict answers. We consider three methods to do this here. 
The first one is to compute inner product between the sum of $s_{Q_L}$ and $s_{V_L}$ and $s_A$, the answer representation defined in Sec.\ref{sec:langfeat}, as
\begin{multline} \label{eq:inner}
(\mbox{score of the answer encoded as $s_A$})\\ = \sigma \Big(s_A^{\top}\, W\, \big(s_{Q_L} + s_{V_L} \big)\Big),
\end{multline}
where  $\sigma$ is the logistic function and $W$ is a learnable weight matrix. The second and third ones are to use a MLP for computing scores for a set of predefined answers, which is a widely used approach in recent studies. The two differ in how to fuse $s_{Q_L}$ and $s_{V_L}$, i.e., summation
\begin{equation} \label{eq:sum}
(\mbox{score of answers}) = \sigma \Big(\text{MLP}\big(s_{Q_L} + s_{V_L}\big)\Big),
\end{equation}
or concatenation
\begin{equation} \label{eq:cat}
(\mbox{score of answers}) = \sigma \Big(\text{MLP}\big(\begin{bmatrix}s_{Q_L}\\ s_{V_L}\end{bmatrix}\big)\Big),
\end{equation}
where $\text{MLP}$ is a two layer MLP having 1024 hidden units  with  ReLU non-linearity. The first one is the most flexible, as it allows us to deal with any answers that are not considered at the time of training the entire network.

\section{Experiments}

\begin{table}[t]
\begin{center}
\caption{Ablation study on each module of DCNs using the validation set of the Open-Ended task (VQA 2.0). * indicates modules employed in the final  model.
}
\label{table:ablation}
\begin{tabular}{|l|c|c|}
\hline
Category & Detail & Accuracy \\
\hline
\hline
Attention direction & 
I $\leftarrow$ Q
 & 60.95 \\
 & 
I $\rightarrow$ Q
 & 62.63 \\
 &
~~I $\leftrightarrow$ Q*
& 62.94 \\
\hline
Memory size ($K$)
 & 1 & 62.53 \\
 & ~~3* & 62.94 \\
 & 5 & 62.83 \\
\hline
Number $(h)$ of
 & 2 & 62.82 \\
parallel attention & ~~4* &  62.94 \\
 maps & 8 & 62.81 \\
\hline
Number $(L)$ of & 1 &  62.43 \\
stacked layers & 2 & 62.82 \\
 & ~~3* & 62.94 \\
 & 4 & 62.67 \\
\hline
Attention in answer & Attention used* &  62.94 \\
prediction layer & Avg of features & 61.63 \\
\hline
Attention in image & Attention used* &  62.94 \\
extraction layer & Only last conv layer & 62.39 \\
\hline
\end{tabular}
\end{center}
\end{table}

\begin{table*}[t]
\begin{center}
\caption{Results of the proposed method along with published results of others on VQA 1.0 in similar conditions (i.e., a single model; trained without an external dataset).}
\label{table:res_vqa}
\begin{tabular}{|c|cccc|cccc|}
\hline
Model & \multicolumn{4}{|c|}{Test-dev} & \multicolumn{4}{|c|}{Test-standard} \\
 & Overall & Other & Number & Yes/No & Overall & Other & Number & Yes/No \\ 
\hline
\hline
VQA team \cite{VQA} & 57.75 & 43.08 & 36.77 & 80.50 & 58.16 & 43.73 & 36.53 & 80.569 \\
SMem \cite{Xu2016} & 57.99 & 43.12 & 37.32 & 80.87 & 58.24 & 43.48 & 37.53 & 80.80 \\
SAN \cite{DBLP:conf/cvpr/YangHGDS16} & 58.70 & 46.10 & 36.60 & 79.30 & 58.90 & - & - & - \\
FDA \cite{DBLP:journals/corr/IlievskiYF16} & 59.24 & 45.77 & 36.16 & 81.14 & 59.54 & - & - & - \\
DNMN \cite{DBLP:conf/naacl/AndreasRDK16} & 59.40 & 45.50 & 38.60 & 81.10 & 59.40 & - & - & - \\
HieCoAtt \cite{DBLP:conf/nips/LuYBP16} & 61.00 & 51.70 & 38.70 & 79.70 & 62.10 & - & - & - \\
RAU \cite{DBLP:journals/corr/NohH16} & 63.30 & 53.00 & 39.00 & 81.90 & 63.20 & 52.80 & 38.20 & 81.70 \\
DAN \cite{DBLP:journals/corr/NamHK16} & 64.30 & 53.90 & 39.10 & 83.00 & 64.20 & 54.00 & 38.10 & 82.80 \\
Strong Baseline \cite{DBLP:journals/corr/KazemiE17} & 64.50 & 55.20 & 39.10 & 82.20 & 64.60 & 55.20 & 39.10 & 82.00 \\
MCB \cite{DBLP:conf/emnlp/FukuiPYRDR16} & 64.70 & 55.60 & 37.60 & 82.50 & - & - & - & - \\
N2NMNs \cite{DBLP:journals/corr/HuARDS17} & 64.90 & - & - & - & - & - & - & - \\
MLAN \cite{multi-level-attention-networks} & 64.60 & 53.70 & 40.20 & 83.80 & 64.80 & 53.70 & 40.90 & 83.70 \\
MLB \cite{Kim2016c} & 65.08 & 54.87 & 38.21 & 84.14 & 65.07 & 54.77 & 37.90 & 84.02 \\
MFB \cite{Yu2017} & 65.90 & 56.20 & 39.80 & 84.00 & 65.80 & 56.30 & 38.90 & 83.80 \\
MF-SIG-T3 \cite{chen2017sva} & 66.00 & 56.37 & 39.34 & 84.33 & 65.88 & 55.89 & 38.94 & 84.42 \\
DCN (\ref{eq:inner}) & 66.43 & 56.23 & 42.37 & 84.75 & 66.39 & 56.23 & 41.81 & 84.53 \\
DCN (\ref{eq:sum}) & {\bf66.89} & 57.31 & {\bf42.35} & {\bf84.61} & {\bf67.02} & {\bf56.98} & {\bf42.34} & {\bf85.04} \\
DCN (\ref{eq:cat}) & 66.83 & {\bf57.44} & 41.66 & 84.48 & 66.66 & 56.83 & 41.27 & 84.61 \\
\hline
\end{tabular}
\end{center}
\end{table*}

\begin{table*}[t]
\begin{center}
\caption{Results of the proposed method along with published results of others on VQA 2.0 in similar conditions (i.e., a single model; trained without an external dataset). DCN(number) indicates the DCN equipped with the prediction layer that uses equation (number) for score computation. *: trained with external datasets. $\ddagger$: the winner of VQA challenge 2017, unpublished.}
\label{table:res_vqa2.0}
\begin{tabular}{|c|cccc|cccc|}
\hline
Model & \multicolumn{4}{|c|}{Test-dev} & \multicolumn{4}{|c|}{Test-standard} \\
 & Overall & Other & Number & Yes/No & Overall & Other & Number & Yes/No\\ 
\hline
\hline
VQA team-Prior \cite{balanced_vqa_v2} & - & - & - & - & 25.98 & 01.17 & 00.36 & 61.20 \\
VQA team-Language only \cite{balanced_vqa_v2} & - & - & - & - & 44.26 & 27.37 & 31.55 & 67.01 \\
VQA team-LSTM+CNN \cite{balanced_vqa_v2} & - & - & - & - & 54.22 & 41.83 & 35.18 & 73.46 \\
MCB \cite{DBLP:conf/emnlp/FukuiPYRDR16} reported in \cite{balanced_vqa_v2} & - & - & - & - & 62.27 & 53.36 & 38.28 & 78.82 \\
MF-SIG-T3 * \cite{chen2017sva} & 64.73 & 55.55 & 42.99 & 81.29 & - & - & - & - \\
Adelaide Model * $\ddagger$ \cite{DBLP:journals/corr/abs-1708-02711} & 62.07 & 52.62 & 39.46 & 79.20 & 62.27 & 52.59 & 39.77 & 79.32 \\
Adelaide + Detector * $\ddagger$
\cite{DBLP:journals/corr/abs-1708-02711} & 65.32 & 56.05 & 44.21 & 81.82 & 65.67 & 56.26 & 43.90 & 82.20 \\
DCN (\ref{eq:inner}) & {\bf66.87} & {\bf57.26} & 46.61 & 83.51 & 66.97 & {\bf57.09} & 46.98 & 83.59 \\
DCN (\ref{eq:sum}) & 66.72 & 56.77 & {\bf46.65} & {\bf83.70} & {\bf67.04} & 56.95 & {\bf47.19} & 83.85 \\
DCN (\ref{eq:cat}) & 66.60 & 56.72 & 46.60 & 83.50 & 67.00 & 56.90 & 46.93 & {\bf83.89} \\
\hline
\end{tabular}
\end{center}
\end{table*}

\begin{figure*}
\centering

\begin{minipage}{1\textwidth}
\centering

\begin{minipage}{0.47\textwidth}
\centering

\begin{minipage}[b]{.45\linewidth}
\includegraphics[width=\linewidth]{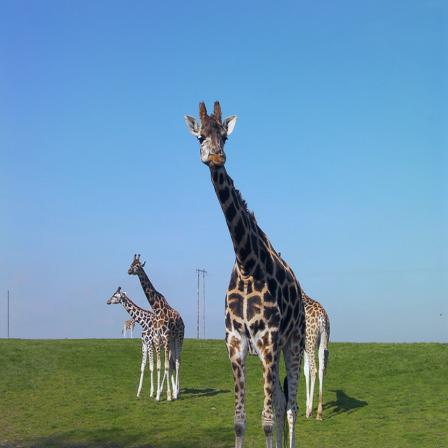}

{\footnotesize \centering What are these animals\par}
\end{minipage}
\begin{minipage}[b]{.45\linewidth}
\includegraphics[width=\linewidth]{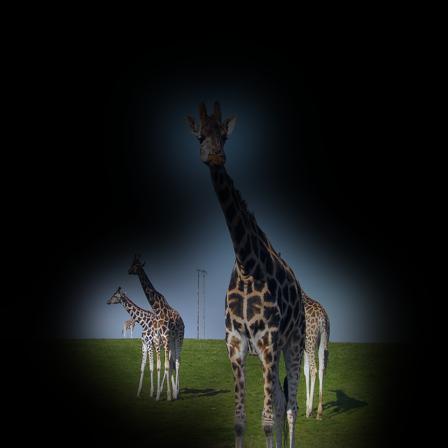}

{\footnotesize \centering {\color[rgb]{1,0.6,0.6}What} {\color[rgb]{1,0.3,0.3}are} {\color[rgb]{1,0.9,0.9}these} {\color[rgb]{1,0,0}animals}\par}
\end{minipage}

{\footnotesize Pred: Giraffes, Ans: Giraffes}
\end{minipage}
\begin{minipage}{0.47\textwidth}
\centering

\begin{minipage}[b]{.45\linewidth}
\includegraphics[width=\linewidth]{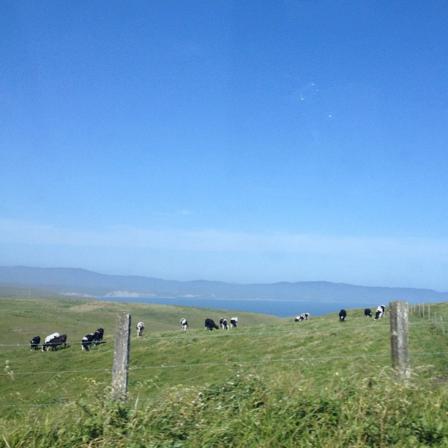}

{\footnotesize \centering What are these animals\par}
\end{minipage}
\begin{minipage}[b]{.45\linewidth}
\includegraphics[width=\linewidth]{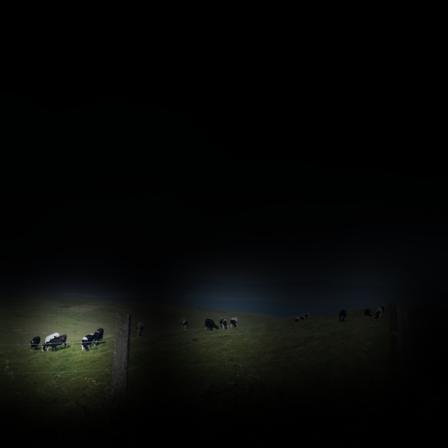}

{\footnotesize \centering {\color[rgb]{1,0.6,0.6}What} {\color[rgb]{1,0.3,0.3}are} {\color[rgb]{1,0.9,0.9}these} {\color[rgb]{1,0,0}animals}\par}
\end{minipage}

{\footnotesize Pred: Cows, Ans: Cows\par}
\end{minipage}
\end{minipage}

\smallskip
\begin{minipage}{1\textwidth}
\centering

\begin{minipage}{0.47\textwidth}
\centering

\begin{minipage}[b]{.45\linewidth}
\includegraphics[width=\linewidth]{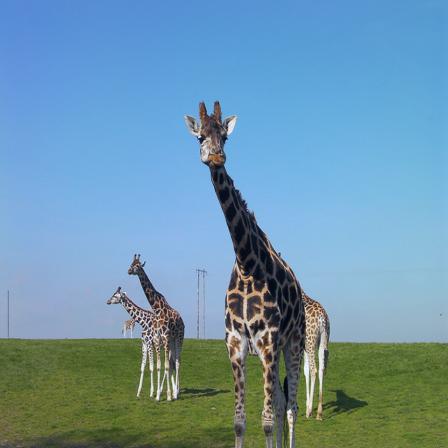}

{\footnotesize \centering Is it cloudy\par}
\end{minipage}
\begin{minipage}[b]{.45\linewidth}
\includegraphics[width=\linewidth]{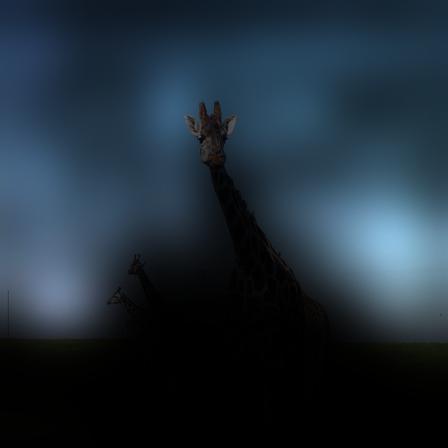}

{\footnotesize \centering {\color[rgb]{1,0.9,0.9}Is} {\color[rgb]{1,0.3,0.3}it} {\color[rgb]{1,0,0}cloudy}\par}
\end{minipage}

{\footnotesize Pred: No, Ans: No\par}
\end{minipage}
\begin{minipage}{0.47\textwidth}
\centering

\begin{minipage}[b]{.45\linewidth}
\includegraphics[width=\linewidth]{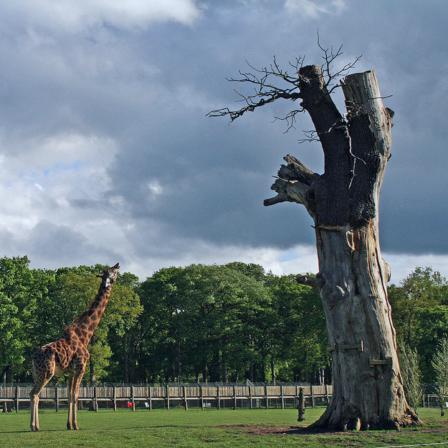}

{\footnotesize \centering Is it cloudy\par}
\end{minipage}
\begin{minipage}[b]{.45\linewidth}
\includegraphics[width=\linewidth]{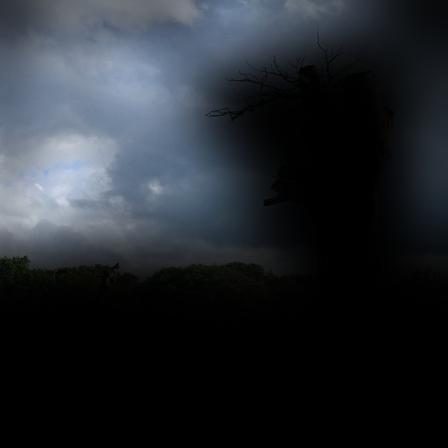}

{\footnotesize \centering {\color[rgb]{1,0.9,0.9}Is} {\color[rgb]{1,0.3,0.3}it} {\color[rgb]{1,0,0}cloudy}\par}
\end{minipage}

{\footnotesize Pred: Yes, Ans: Yes\par}
\end{minipage}
\end{minipage}

\caption{Typical examples of attended image regions and question words for complementary image-question pairs from VQA 2.0 dataset. Each row contains visualization for two pairs of the same question but different images and answers. The original image and question are shown along with their attention maps generated in the answer prediction layer. The brightness of image pixels and redness of words indicate the attention weights.}
\label{fig:visualization}
\end{figure*}

In this section, we present results of the experiments conducted to evaluate the proposed method. 

\subsection{Datasets}

We used two most popular datasets, VQA \cite{VQA} and VQA 2.0 \cite{balanced_vqa_v2}, for our experiments.   
VQA (also known as  VQA 1.0) contains 
human-annotated question-answer pairs on 204,721 images from Microsoft COCO dataset \cite{MSCOCO}. There are three predefined  splits of questions, {\em train}, {\em val} and {\em test} or {\em test-standard}, which consist of 248,349, 121,512, and 244,302 questions, respectively. There is also a $25\%$ subset of the test-standard set referred to as {\em test-dev}.
All the questions are categorized into three types: yes/no, number, and other. Each question has 10 free-response answers. VQA 2.0 is an updated version of VQA 1.0 and is the largest as of now. Compared with VQA 1.0, it contains more samples (443,757 {\em train}, 214,354 {\em val}, and 447,793 {\em test} questions), and is more balanced in term of language bias.
We evaluate our models on the challenging Open-Ended task of both datasets.

As in \cite{DBLP:journals/corr/abs-1708-02711}, we choose correct answers appearing more than $5$ times for VQA and $8$ times for VQA 2.0 to form the set of candidate answers. 
Following previous studies, we train our network on train + val splits and report the test-dev and test-standard results from the VQA evaluation server (except for the ablation test shown below). We use the evaluation protocol of \cite{VQA} in all the experiments.

\subsection{Experimental Setup}

For both of the datasets, we use the Adam optimizer with 
the parameters $\alpha=0.001$, 
$\beta_1 = 0.9$, and $\beta_2 = 0.99$. During the training procedure, we make the learning rate $(\alpha)$ decay at every 4 epochs for VQA and 7 epochs for VQA 2.0 with an exponential rate of 0.5. All models are trained up to 16 and 21 epochs on VQA and VQA 2.0, respectively. To prevent overfitting, dropouts \cite{JMLR:v15:srivastava14a} are used after each fully connected layers with a dropout ratio $p = 0.3$ and after the LSTM with a dropout ratio $p = 0.1$. The batch size is set to 160 and 320  for VQA and VQA 2.0. We set the dimension $d$ of the feature space in the dense co-attention layers (equivalently, the size of its hidden layers) to be 1024.

\subsection{Ablation Study}

The architecture of the proposed DCN is composed of multiple modules. To evaluate the contribution of each module to final prediction accuracy, we conducted ablation tests. Using VQA 2.0, we evaluated several versions of DCNs with different parameters and settings by  training them on the train split and calculating its performance on the val split. The results are shown in Table \ref{table:ablation}.

The first block of the table shows the effects of image-question co-attention. The numbers are performances obtained by a DCN with only question-guided attention on images  $(I\leftarrow Q)$, with only image-guided attention on question words $(I\rightarrow Q)$, and the standard DCN with co-attention $(I\leftrightarrow Q)$. The single-direction variants generates only attention in either side of the two paths in the dense co-attention layer; the rest of the computations remain the same. The network with co-attention performs the best, verifying the effectiveness of our co-attention implementation. 

The second block of the table shows the impacts of $K$, which is the row size of $M_{Q_l}$ and $M_{V_l}$ that are used for augmenting $Q_l$ and $V_l$. This augmentation is originally introduced to be able to deal with ``nowhere to attend", which can be implemented by $K=1$ \cite{XiongMS16}. However, we found that the use of $K>1$ improves performance to a certain extent, which we think is because $M_{Q_l}$ and $M_{V_l}$ work as external memory that can be used through attention mechanism \cite{DBLP:journals/corr/GravesWD14}. 
As shown in the table, $K=3$ yields the best performance.

The third and fourth blocks of the table show choices of the number $h$ of parallel attention maps and $L$ of stacked layers. The best result was obtained for $h=4$ and $L = 3$.

The last two blocks of the table show effects of the use of attention in the answer prediction layer and the image extraction layer; the use of attention improves accuracy by about $1.3\%$ and $0.5\%$, respectively.

\subsection{Comparison with Existing Methods}

Table \ref{table:res_vqa} shows the performance of our method on VQA 1.0 along with published results of others. The entries `DCN ($n$)' indicate which method for score computation is employed from (\ref{eq:inner})-(\ref{eq:cat}). It is seen from the table that our method outperforms the best published result (MF-SIG-T3) by a large margin of $0.9\% \sim 1.1\%$ on both test-dev and test-standard sets. Furthermore, the improvements can be seen in all of the entries ({\em Other} with $1.1\%$, {\em Number} with $3.4\%$, {\em Yes/No} with $0.6\%$ on test-standard set) implying the capacity of DCNs to model multiple types of complex relations between question-image pairs. Notably, we achieve significant improvements of $3.0\%$ and $3.4\%$ for the question type {\em Number} on test-dev and test-standard sets, respectively.

Table 
\ref{table:res_vqa}
also shows the performances of DCNs with a different answer prediction layer that uses (\ref{eq:inner}), (\ref{eq:sum}), and (\ref{eq:cat}) for score computation. It is seen that (\ref{eq:sum}) shows at least comparable performance to the others and even attains the best performance of $67.02\%$ in test-standard set.

Table \ref{table:res_vqa2.0} shows comparisons of our method to previous published results on VQA 2.0 and also that of the winner of VQA 2.0 Challenge 2017 
in both test-dev and test-standard sets.
It is observed in Table \ref{table:res_vqa2.0} that our approach outperforms the state-of-the-art published method (MF-SIG-T3) by a large margin of $2.1\%$ on test-dev set, even though the MF-SIG-T3 model was trained with VQA 2.0 and an augmented dataset (Visual Genome \cite{Krishna:2017:VGC:3088990.3089101}). It is noted that the improvements are seen in all the question types ({\em Other} with $1.71\%$, {\em Number} with $3.66\%$, and {\em Yes/No} with $2.41\%$). Comparing our DCN with the winner of VQA 2.0 Challenge 2017, Adelaide model. Our best DCN  (\ref{eq:sum}) delivers $1.5\%$ and $1.37\%$ improvements in every question types over the Adelaide+Detector on test-dev and test-standard, respectively. It is worth to point out that the winner method uses a detector (Region Proposal Network) trained on annotated regions of Visual Genome dataset \cite{Krishna:2017:VGC:3088990.3089101} to extract visual features; and that the model is trained using also an external dataset, i.e., the Visual Genome question answering dataset.

It should also be noted that while achieving the best performance in VQA dataset, the size of the DCNs (i.e., the number of parameters) is comparable or even smaller than the former state-of-the-art methods, as shown in Table \ref{table:modelsize}.
\begin{table}[t]
\begin{center}
\caption{Model sizes of DCNs and several bilinear fusion methods. The numbers include the parameters of LSTM networks and exclude those of ResNets.}
\label{table:modelsize}
\begin{tabular}{|l|c|}
\hline
Model & No. params \\
\hline
\hline
MCB \cite{DBLP:conf/emnlp/FukuiPYRDR16} & 63M \\
MLB \cite{Kim2016c} & 25M \\
MFB \cite{Yu2017} & 46M \\
DCN (\ref{eq:cat}) & 32M \\
DCN (\ref{eq:sum}) & 31M \\
DCN (\ref{eq:inner}) & 28M \\
\hline
\end{tabular}
\end{center}
\end{table}

\subsection{Qualitative Evaluation}

Complementary image-question pairs are available in VQA 2.0 \cite{balanced_vqa_v2}, which are pairs of the same question and different images with different answers. To understand the behaviour of the trained DCN, we visualize attention maps that the DCN generates for some of the complementary image-question pairs. Specifically, we show multiplication of an input image and question with their attention maps $\alpha^V_1,\ldots,\alpha^V_T$ and $\alpha^Q_1,\ldots,\alpha^Q_N$ (defined in Sec.\ref{sec:anspred}) generated in the answer prediction layer. A typical example is shown in Fig.\ref{fig:visualization}. Each row shows the results for two pairs of the same question and different images, from which we can observe that the DCN is able to look at right regions to find the correct answers. Then, the first column shows the results for two pairs of the same image and different questions. It is observed that the DCN focuses on relevant image regions and question words to produce answers correctly.
More visualization results including failure cases are provided in the supplementary material.

\section{Conclusion}

In this paper, we present a novel network architecture for VQA named the dense co-attention network. 
The core of the network is the dense co-attention layer, which is designed to enable improved fusion of visual and language representations by considering dense symmetric interactions between the input image and question. The layer can be stacked to perform multi-step image-question interactions. The layer stack combined with the initial feature extraction step and the final answer prediction layer, both of which have their own attention mechanisms, form the dense co-attention network. The experimental results on two datasets, VQA and VQA 2.0, confirm the effectiveness of the proposed architecture. 

\section*{Acknowledgement}

{\normalsize
This work was partly supported by JSPS KAKENHI Grant Number JP15H05919, JST CREST Grant Number JPMJCR14D1, Council for Science, Technology and Innovation (CSTI), Cross-ministerial Strategic Innovation Promotion Program (Infrastructure Maintenance, Renovation and Management ), and the ImPACT Program “Tough Robotics Challenge” of the Council for Science, Technology, and Innovation (Cabinet Office, Government of Japan).
}

{\small
\bibliographystyle{ieee}
\bibliography{egbib}
}

\onecolumn
\appendix

This document contains: more details of our experimental setups (Sec.\ref{sec:detail}), the evaluation of effects in the employment of Contextualized Word Vectors (Sec.\ref{sec:cove}), more visualization of attention maps generated in the answer prediction layer including failure cases (Sec.\ref{sec:attn_pred}), and 
an analysis of the attention mechanism employed in the image feature extraction
(Sec.\ref{sec:result_extrac}).

\section{More Details of the Experimental Setups}
\label{sec:detail}

In our experiments, images and questions are preprocessed as follows. All the images were resized to $448 \times 448$ before feeding into the CNN. All the questions were tokenized using Python Natural Language Toolkit (nltk). We used the vocabulary provided by the CommonCrawl-840B Glove model for English word vectors \cite{pennington2014glove}, and set out-of-vocabulary words to {\it unk}. As mentioned in the main paper, we chose the correct answer appearing more than 5 times (= 3,014 answers) for VQA 1.0, and 8 times (= 3,113 answers) for VQA 2.0 as in \cite{DBLP:journals/corr/abs-1708-02711}. We capped the maximum length of questions at 14 words and then performed dynamic unrolling for each question to allow for questions of different lengths.

Throughout the experiments, we used three-layer DCNs, that is, DCNs with three dense co-attention layers (L = 3). This number of layers were chosen based on our preliminary experiments. 
The Bi-LSTM was initialized following the recommendation in \cite{EmpiricalRNN} and all the other parameters were initialized as suggested by Glorot \etal. In the training procedure, the ADAM \cite{adam} optimizer was used to train our model for 16 and 21 epochs on VQA and VQA 2.0 with batch size of 160 and 320, respectively;
weight decay with rate of 0.0001 was added. We used exponential decay to gradually decrease the learning rate as
\[\alpha_{step} = 0.5^{\frac{\text{epochs}}{\text{decay epochs}}}\alpha, \]
where the initial learning rate $\alpha$ was set to $\alpha=0.001$, and the decay epochs was set to 4 and 7 epochs for VQA and VQA 2.0 in turn; we set $\beta_1=0.9$, and $\beta_2=0.99$.

\begin{figure}
\centering

\begin{minipage}{1\textwidth}
\centering

\begin{minipage}{0.47\textwidth}
\centering

\includegraphics[width=\linewidth]{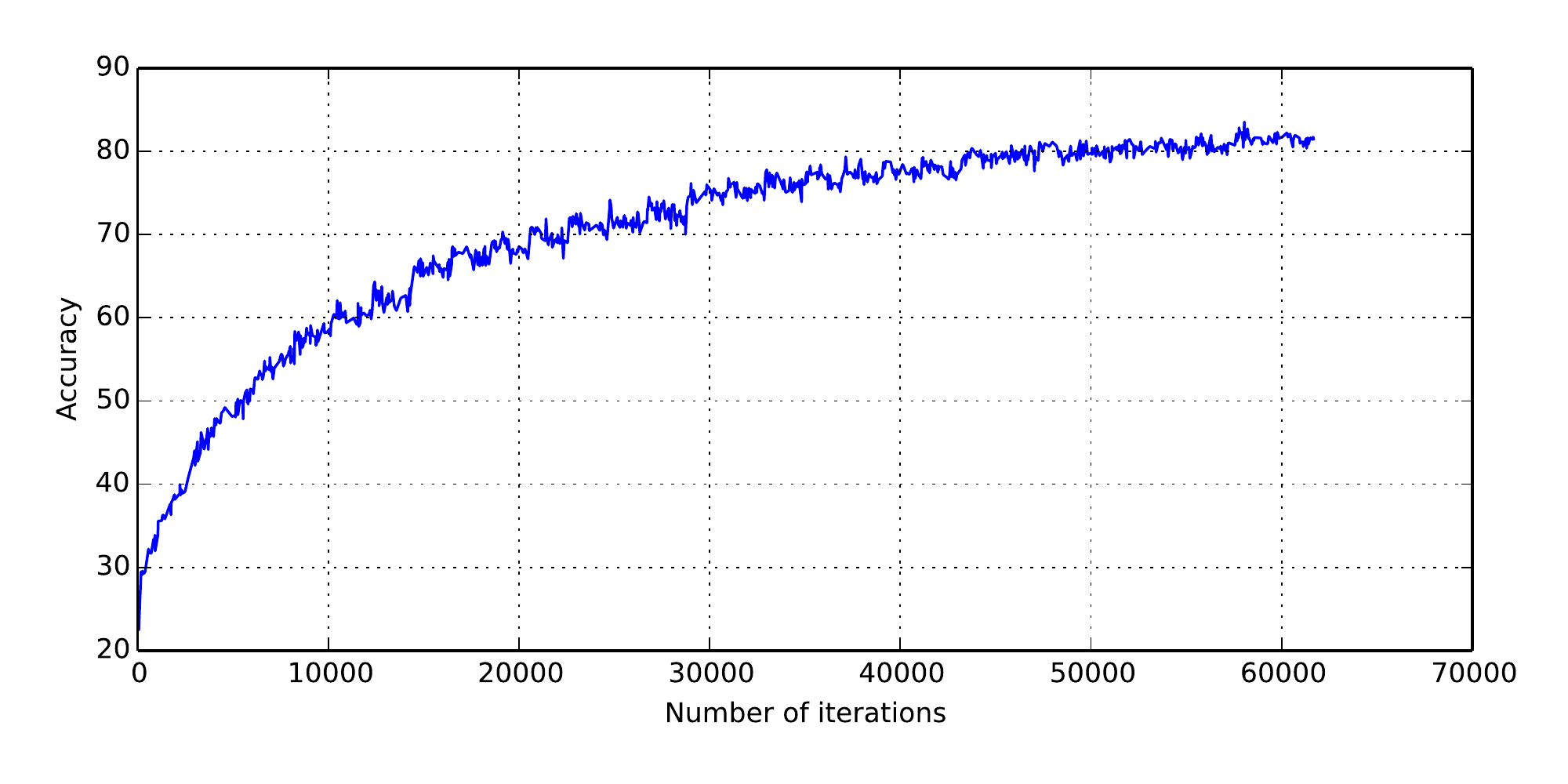}

\end{minipage}
\begin{minipage}{0.47\textwidth}
\centering

\includegraphics[width=\linewidth]{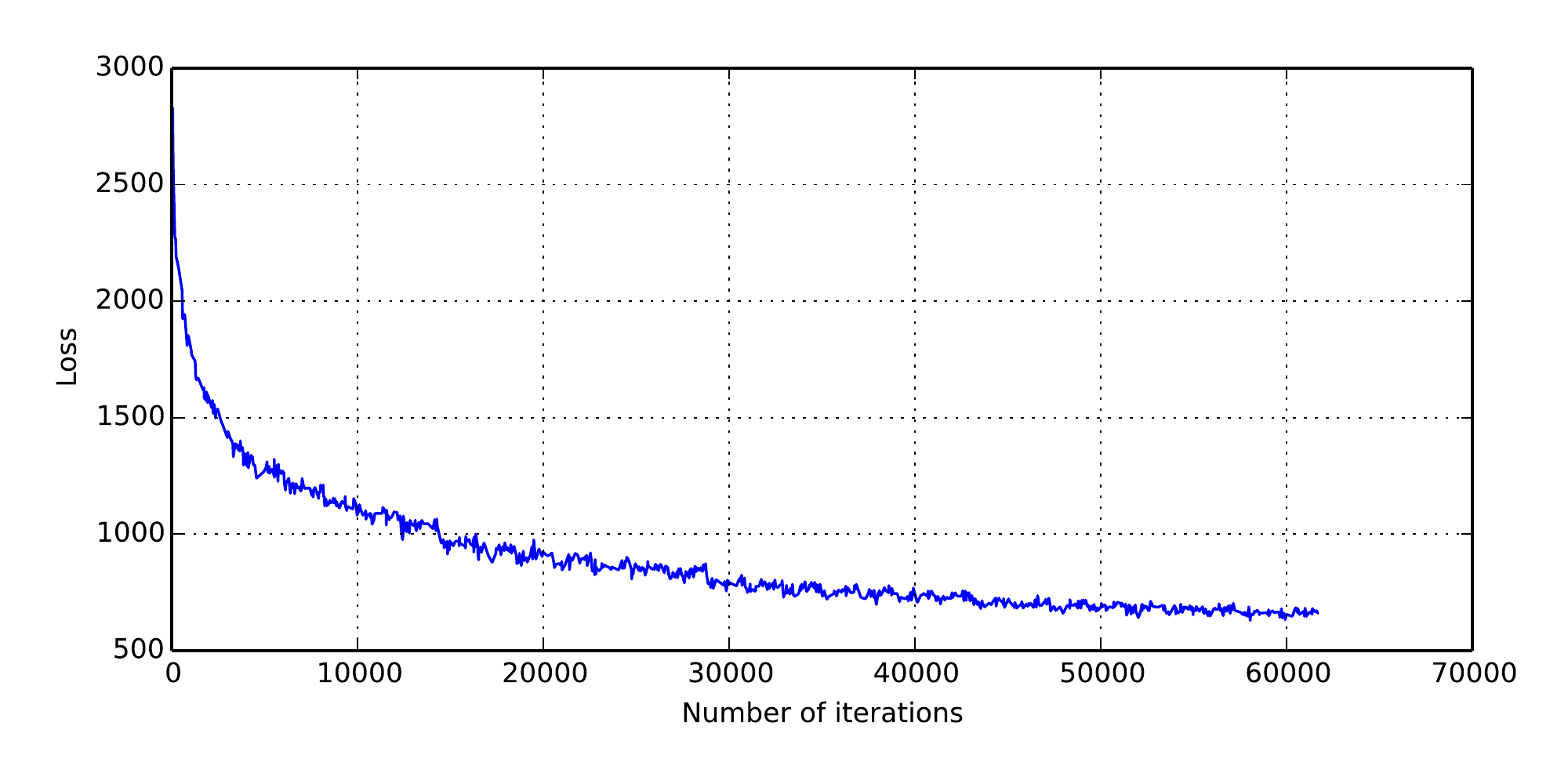}

\end{minipage}
\end{minipage}
\setcounter{figure}{4}
\caption{Learning curves for DCN.}
\label{fig:training_curve}
\end{figure}

\clearpage
\section{Effects of the Employment of Contextualized Word Vectors}
\label{sec:cove}

To extract word features from input questions, some of the previous studies \cite{kim2016b, Kim2016c, chen2017sva, Ben-younesCCT17} employed 
pretrained RNNs (specifically, GRU networks pre-trained with Skip-thought) \cite{Kiros:2015:SV:2969442.2969607}. In this study, we initially pursued a similar approach; we perform fine-tuning of a pretrained LSTM, specifically a two-layer Bi-LSTM trained as a CoVe (Context Vector) encoder \cite{McCann2017LearnedIT}. Conducting comparative experiments, we eventually employ a single-layer Bi-LSTM with random initialization, as explained in the main paper. We report here the results of the experiments. 


Table \ref{table:cove} shows the performances of DCNs with the CoVe-pretrained Bi-LSTM and with the randomly initialized Bi-LSTM. Note that the former is a two-layer model and the later has only one layer. Here, the VQA 2.0 test-dev dataset was used. It is observed that for DCNs with the answer prediction layer of (16), the one with the CoVe-pretrained model performs slightly better than the one with the randomly initialized model, but their differences are small. For DCNs with the answer prediction layers of (17) and (18), the one with the randomly initialized model performs better with a less number of parameters.

It should be noted, however, that the employment of CoVe-pretrained models, together with the answer prediction layer of (16), enables to compute meaningful answer representation $(s_A)$ for answers that have not been seen before, i.e., those that are not included in training data. Table \ref{table:mc} shows the results of DCN (16) with the CoVe-pretrained model for {\em Multiple Choice} answers, which include a lot of unseen answers. 
This is not the case with DCNs (17) and (18) that compute scores of a fixed set of predetermined answers--- the common approach of most of the recent studies.

\begin{table}[ht!]
\begin{center}
\setcounter{table}{4}
\caption{Performances of DCNs with the CoVe-pretrained LSTM and with the randomly initialized LSTM on the VQA 2.0 test-dev set. 
}
\label{table:cove}
\begin{tabular}{|c|c|c|c|c|c|c|}
\hline
Model & Overall & Other & Number & Yes/No & No. params\\
\hline
\hline
DCN (16) + CoVe & 67.06 & 57.44 & 46.91 & 83.69 & 31M \\
DCN (16) & 66.87 & 57.26 & 46.61 & 83.51 & 28M \\
\hline
DCN (17) + CoVe & 66.21 & 56.71 & 46.01 & 82.72 & 34M \\
DCN (17) & 66.72 & 56.77 & 46.65 & 83.70 & 31M \\
\hline
DCN (18) + CoVe & 66.31 & 56.62 & 45.78 & 83.14 & 35M \\
DCN (18) & 66.60 & 56.72 & 46.60 & 83.50 & 32M \\
\hline
\end{tabular}
\end{center}
\end{table}

\begin{table}[ht!]
\begin{center}
\setcounter{table}{5}
\caption{Effectiveness of DCN (16) + CoVe-pretrained LSTM on {\em Multiple Choice} answers.}
\label{table:mc}
\begin{tabular}{|c|cccc|cccc|}
\hline
Model & \multicolumn{4}{|c|}{Test-dev} & \multicolumn{4}{|c|}{Test-std} \\
& Overall & Other & Number & Yes/No & Overall & Other & Number & Yes/No \\ 
\hline
\hline
DCN (16) + CoVe & 71.37 & 66.10 & 45.48 & 84.39 & 71.20 & 65.93 & 44.13 & 84.23 \\
\hline
\end{tabular}
\end{center}
\end{table}

\clearpage
\section{Visualization of Attention Maps in the Answer Prediction Layer}
\label{sec:attn_pred}

We have shown a few examples of attention maps generated in the answer prediction layer of our DCNs in Fig.4 of the main paper. We show here more examples for success cases (Sec.\ref{sec:attn_pred_suc}) and also for failure cases (Sec.\ref{sec:attn_pred_fail}). 

\subsection{Success Cases}
\label{sec:attn_pred_suc}

We consider the visualization of complementary pairs to analyze the behaviour of our DCNs. Each row shows a complementary pair having the same question and different images. It can be seen from the examples shown below that the image and question attention maps are generated appropriately for most of success cases. 

\begin{figure*}[hpt!]
\centering

\bigskip
\bigskip
\bigskip
\begin{minipage}{1\textwidth}
\centering

\fcolorbox{green}{white}{
\begin{minipage}{.45\textwidth}
\centering

\begin{minipage}[b]{.45\linewidth}
\includegraphics[width=\linewidth]{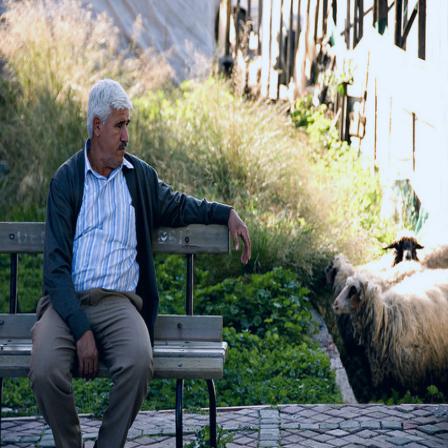}

{\footnotesize \centering What is he sitting on \par}
\end{minipage}
\begin{minipage}[b]{.45\linewidth}
\includegraphics[width=\linewidth]{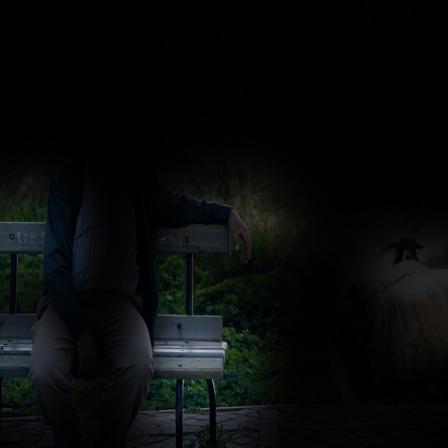}

{\footnotesize \centering {\color[rgb]{1,0.2,0.2}What} {\color[rgb]{1,0.5,0.5}is} {\color[rgb]{1,0.5,0.5}he} {\color[rgb]{1,0.6,0.6}sitting} {\color[rgb]{1,0.9,0.9}on} \par}
\end{minipage}

{\footnotesize Pred: Bench, Ans: Bench\par}
\end{minipage}
}
\fcolorbox{green}{white}{
\begin{minipage}{.45\textwidth}
\centering

\begin{minipage}[b]{.45\linewidth}
\includegraphics[width=\linewidth]{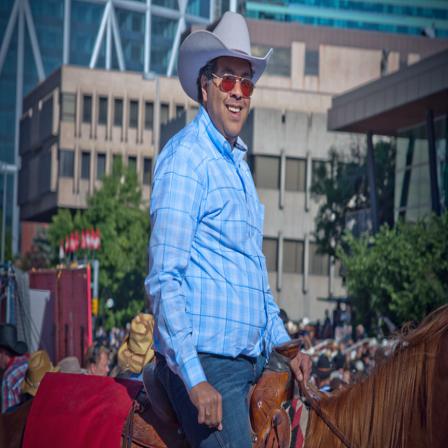}

{\footnotesize \centering What is he sitting on \par}
\end{minipage}
\begin{minipage}[b]{.45\linewidth}
\includegraphics[width=\linewidth]{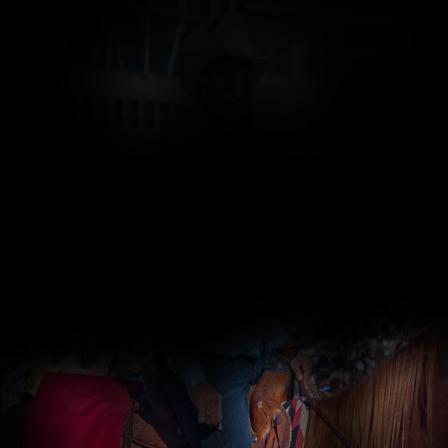}

{\footnotesize \centering {\color[rgb]{1,0.2,0.2}What} {\color[rgb]{1,0.2,0.2}is} {\color[rgb]{1,0.4,0.4}he} {\color[rgb]{1,0.6,0.6}sitting} {\color[rgb]{1,0.9,0.9}on} \par}
\end{minipage}

{\footnotesize Pred: Horse, Ans: Horse\par}
\end{minipage}
}
\end{minipage}

\smallskip
\begin{minipage}{1\textwidth}
\centering

\fcolorbox{green}{white}{
\begin{minipage}{.45\textwidth}
\centering

\begin{minipage}[b]{.45\linewidth}
\includegraphics[width=\linewidth]{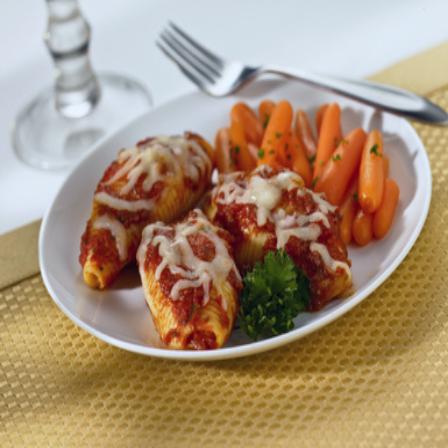}

{\footnotesize \centering What type of meal is this\par}
\end{minipage}
\begin{minipage}[b]{.45\linewidth}
\includegraphics[width=\linewidth]{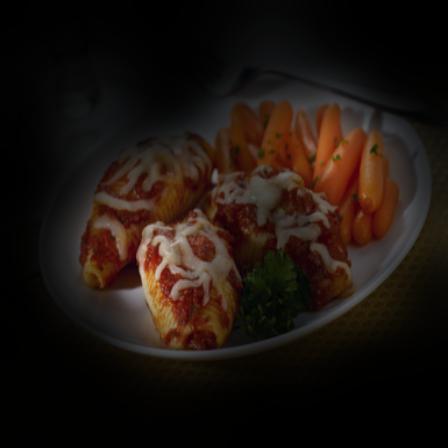}

{\footnotesize \centering {\color[rgb]{1,0.6,0.6}What type} {\color[rgb]{1,0.9,0.9}of} {\color[rgb]{1,0,0}meal} {\color[rgb]{1,0.9,0.9}is this} \par}
\end{minipage}

{\footnotesize Pred: Dinner, Ans: Dinner\par}
\end{minipage}
}
\fcolorbox{green}{white}{
\begin{minipage}{.45\textwidth}
\centering

\begin{minipage}[b]{.45\linewidth}
\includegraphics[width=\linewidth]{correct_pairs/pair13b_pic.png}

{\footnotesize \centering  What type of meal is this\par}
\end{minipage}
\begin{minipage}[b]{.45\linewidth}
\includegraphics[width=\linewidth]{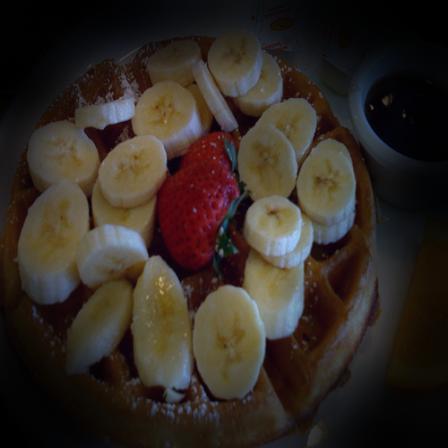}

{\footnotesize \centering {\color[rgb]{1,0.9,0.9}What} {\color[rgb]{1,0.6,0.6}type} {\color[rgb]{1,0.9,0.9}of} {\color[rgb]{1,0,0}meal} {\color[rgb]{1,0.9,0.9}is this} \par}
\end{minipage}

{\footnotesize Pred: Breakfast, Ans: Breakfast\par}
\end{minipage}
}
\end{minipage}

\smallskip
\begin{minipage}{1\textwidth}
\centering

\fcolorbox{green}{white}{
\begin{minipage}{.45\textwidth}
\centering

\begin{minipage}[b]{.45\linewidth}
\includegraphics[width=\linewidth]{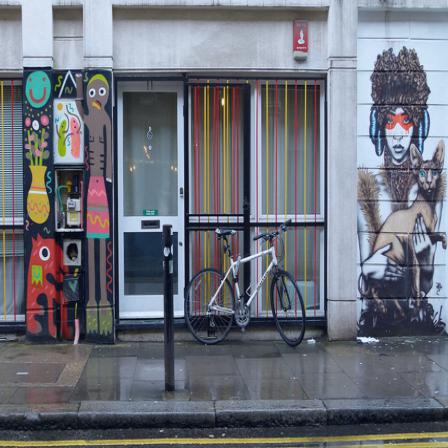}

{\footnotesize \centering Is there a graffiti on the wall \par}
\end{minipage}
\begin{minipage}[b]{.45\linewidth}
\includegraphics[width=\linewidth]{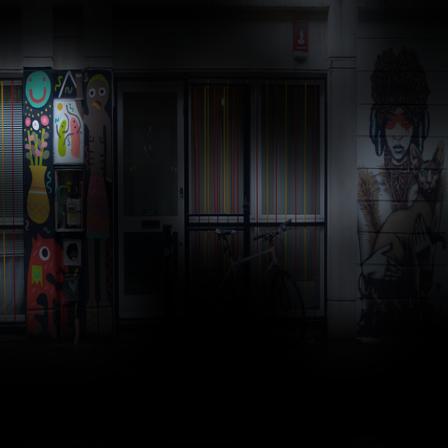}

{\footnotesize \centering {\color[rgb]{1,0.5,0.5}Is} {\color[rgb]{1,0,0}there} {\color[rgb]{1,0.5,0.5}a} {\color[rgb]{1,0.2,0.2}graffiti} {\color[rgb]{1,0.9,0.9}on the} {\color[rgb]{1,0.5,0.5}wall} \par}
\end{minipage}

{\footnotesize Pred: Yes, Ans: Yes\par}
\end{minipage}
}
\fcolorbox{green}{white}{
\begin{minipage}{.45\textwidth}
\centering

\begin{minipage}[b]{.45\linewidth}
\includegraphics[width=\linewidth]{correct_pairs/pair1b_pic.png}

{\footnotesize \centering Is there a graffiti on the wall \par}
\end{minipage}
\begin{minipage}[b]{.45\linewidth}
\includegraphics[width=\linewidth]{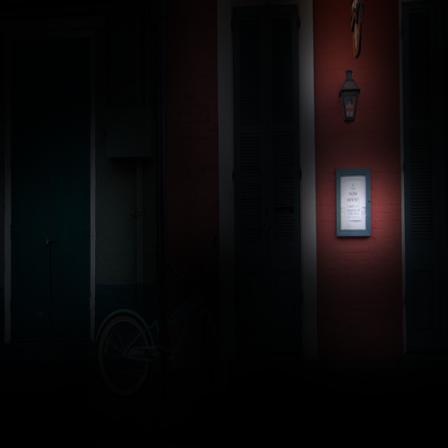}

{\footnotesize \centering {\color[rgb]{1,0.5,0.5}Is} {\color[rgb]{1,0,0}there} {\color[rgb]{1,0.5,0.5}a} {\color[rgb]{1,0.2,0.2}graffiti} {\color[rgb]{1,0.9,0.9}on the} {\color[rgb]{1,0.5,0.5}wall} \par}
\end{minipage}

{\footnotesize Pred: No, Ans: No\par}
\end{minipage}
}
\end{minipage}

\end{figure*}

\begin{figure*}[hpt!]
\centering

\smallskip
\begin{minipage}{1\textwidth}
\centering

\fcolorbox{green}{white}{
\begin{minipage}{.45\textwidth}
\centering

\begin{minipage}[b]{.45\linewidth}
\includegraphics[width=\linewidth]{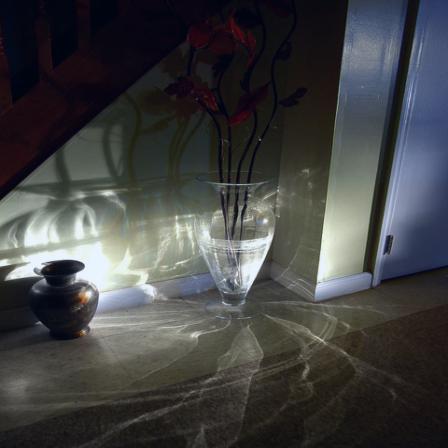}

{\footnotesize \centering How many vases are in the photo \par}
\end{minipage}
\begin{minipage}[b]{.45\linewidth}
\includegraphics[width=\linewidth]{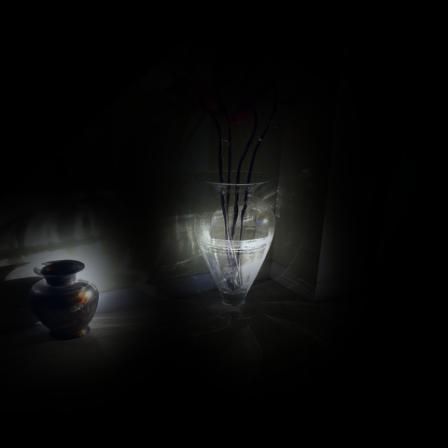}

{\footnotesize \centering {\color[rgb]{1,0.6,0.6}How} {\color[rgb]{1,0,0}many} {\color[rgb]{1,0.9,0.9}vases are in the photo} \par}
\end{minipage}

{\footnotesize Pred: 2, Ans: 2\par}
\end{minipage}
}
\fcolorbox{green}{white}{
\begin{minipage}{.45\textwidth}
\centering

\begin{minipage}[b]{.45\linewidth}
\includegraphics[width=\linewidth]{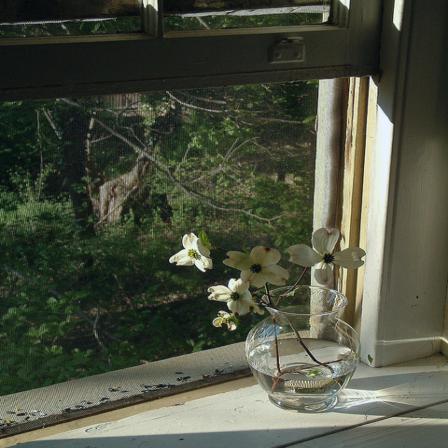}

{\footnotesize \centering How many vases are in the photo \par}
\end{minipage}
\begin{minipage}[b]{.45\linewidth}
\includegraphics[width=\linewidth]{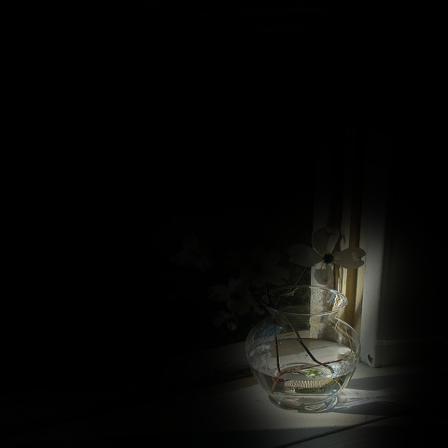}

{\footnotesize \centering {\color[rgb]{1,0.6,0.6}How} {\color[rgb]{1,0,0}many} {\color[rgb]{1,0.9,0.9}vases} {\color[rgb]{1,0.6,0.6}are} {\color[rgb]{1,0.9,0.9}in the photo} \par}
\end{minipage}

{\footnotesize Pred: 1, Ans: 1\par}
\end{minipage}
}
\end{minipage}

\smallskip
\begin{minipage}{1\textwidth}
\centering

\fcolorbox{green}{white}{
\begin{minipage}{.45\textwidth}
\centering

\begin{minipage}[b]{.45\linewidth}
\includegraphics[width=\linewidth]{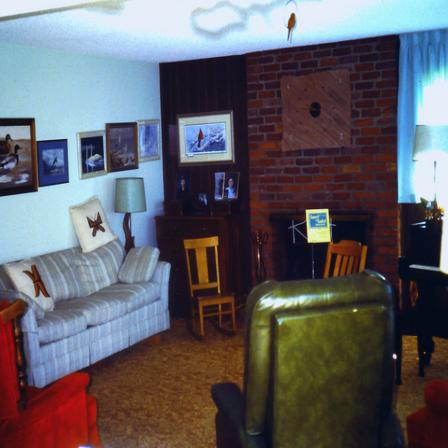}

{\footnotesize \centering What is the darker wall made of \par}
\end{minipage}
\begin{minipage}[b]{.45\linewidth}
\includegraphics[width=\linewidth]{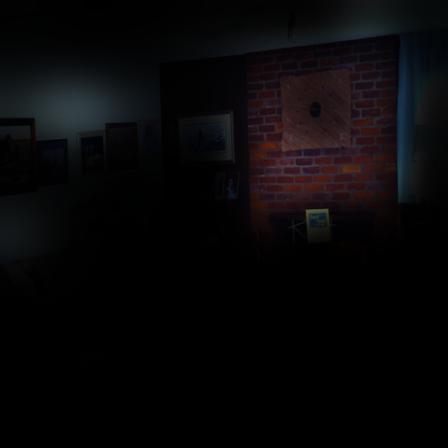}

{\footnotesize \centering {\color[rgb]{1,0.2,0.2}What} {\color[rgb]{1,0.4,0.4}is} {\color[rgb]{1,0.6,0.6}the} {\color[rgb]{1,0.3,0.3}darker wall} {\color[rgb]{1,0.6,0.6} made} {\color[rgb]{1,0.7,0.7}of} \par}
\end{minipage}

{\footnotesize Pred: Brick, Ans: Brick\par}
\end{minipage}
}
\fcolorbox{green}{white}{
\begin{minipage}{.45\textwidth}
\centering

\begin{minipage}[b]{.45\linewidth}
\includegraphics[width=\linewidth]{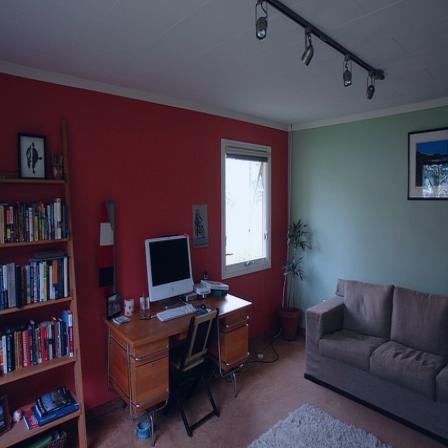}

{\footnotesize \centering What is the darker wall made of \par}
\end{minipage}
\begin{minipage}[b]{.45\linewidth}
\includegraphics[width=\linewidth]{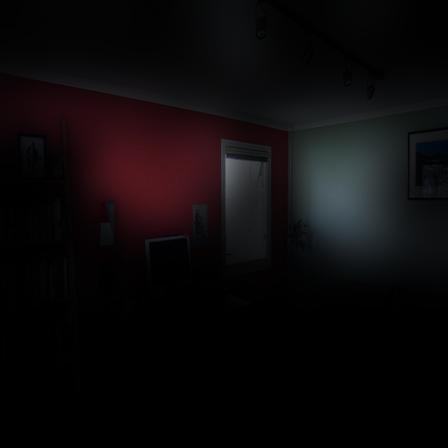}

{\footnotesize \centering {\color[rgb]{1,0.2,0.2}What is} {\color[rgb]{1,0.5,0.5}the} {\color[rgb]{1,0.3,0.3}darker wall} {\color[rgb]{1,0.5,0.5}made} {\color[rgb]{1,0.7,0.7}of} \par}
\end{minipage}

{\footnotesize Pred: Drywall, Ans: Drywall\par}
\end{minipage}
}
\end{minipage}

\smallskip
\begin{minipage}{1\textwidth}
\centering

\fcolorbox{green}{white}{
\begin{minipage}{.45\textwidth}
\centering

\begin{minipage}[b]{.45\linewidth}
\includegraphics[width=\linewidth]{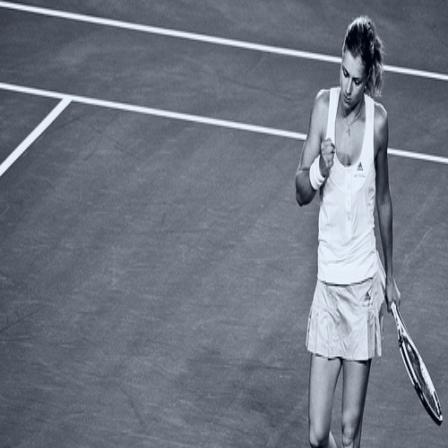}

{\footnotesize \centering What sport is this woman playing \par}
\end{minipage}
\begin{minipage}[b]{.45\linewidth}
\includegraphics[width=\linewidth]{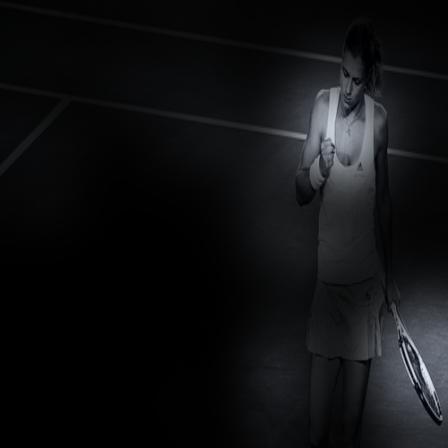}

{\footnotesize \centering {\color[rgb]{1,0.3,0.3}What} {\color[rgb]{1,0,0}sport} {\color[rgb]{1,0.7,0.7}is} {\color[rgb]{1,0.9,0.9}this woman} {\color[rgb]{1,0.5,0.5}playing} \par}
\end{minipage}

{\footnotesize Pred: Tennis, Ans: Tennis\par}
\end{minipage}
}
\fcolorbox{green}{white}{
\begin{minipage}{.45\textwidth}
\centering

\begin{minipage}[b]{.45\linewidth}
\includegraphics[width=\linewidth]{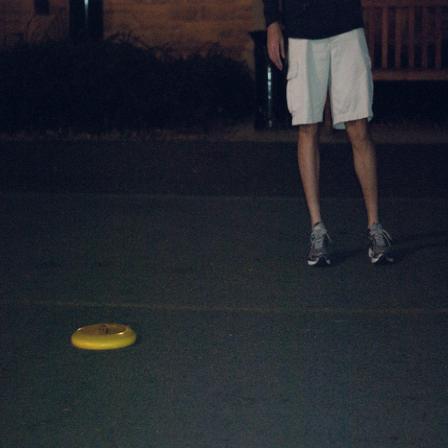}

{\footnotesize \centering What sport is this woman playing \par}
\end{minipage}
\begin{minipage}[b]{.45\linewidth}
\includegraphics[width=\linewidth]{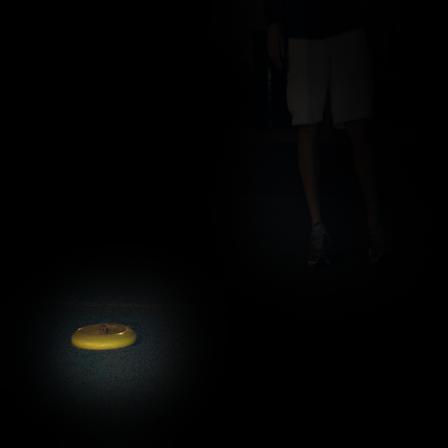}

{\footnotesize \centering {\color[rgb]{1,0.4,0.4}What} {\color[rgb]{1,0,0}sport} {\color[rgb]{1,0.7,0.7}is} {\color[rgb]{1,0.5,0.5}this woman} {\color[rgb]{1,0.5,0.5}playing} \par}
\end{minipage}

{\footnotesize Pred: Frisbee, Ans: Frisbee\par}
\end{minipage}
}
\end{minipage}

\smallskip
\begin{minipage}{1\textwidth}
\centering

\fcolorbox{green}{white}{
\begin{minipage}{.45\textwidth}
\centering

\begin{minipage}[b]{.45\linewidth}
\includegraphics[width=\linewidth]{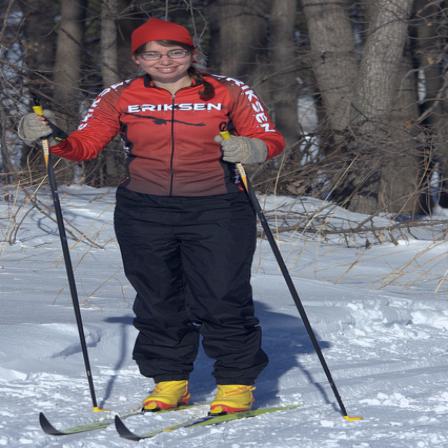}

{\footnotesize \centering What color are the skiers shoes\par}
\end{minipage}
\begin{minipage}[b]{.45\linewidth}
\includegraphics[width=\linewidth]{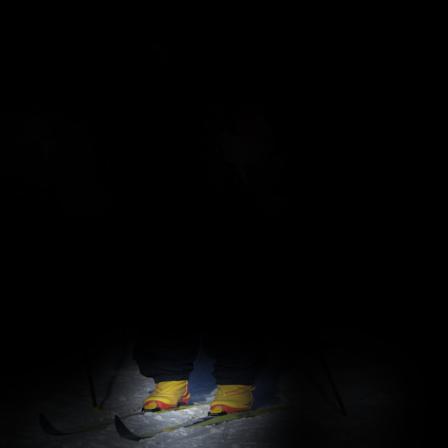}

{\footnotesize \centering {\color[rgb]{1,0.4,0.4}What} {\color[rgb]{1,0,0}color} {\color[rgb]{1,0.9,0.9}are} {\color[rgb]{1,0.5,0.5}the} {\color[rgb]{1,0.9,0.9}skiers} {\color[rgb]{1,0.2,0.2}shoes} \par}
\end{minipage}

{\footnotesize Pred: Yellow, Ans: Yellow\par}
\end{minipage}
}
\fcolorbox{green}{white}{
\begin{minipage}{.45\textwidth}
\centering

\begin{minipage}[b]{.45\linewidth}
\includegraphics[width=\linewidth]{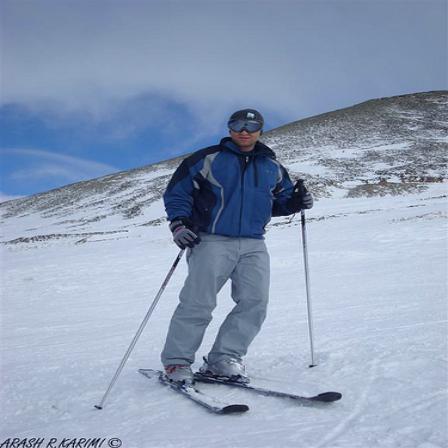}

{\footnotesize \centering What color are the skiers shoes\par}
\end{minipage}
\begin{minipage}[b]{.45\linewidth}
\includegraphics[width=\linewidth]{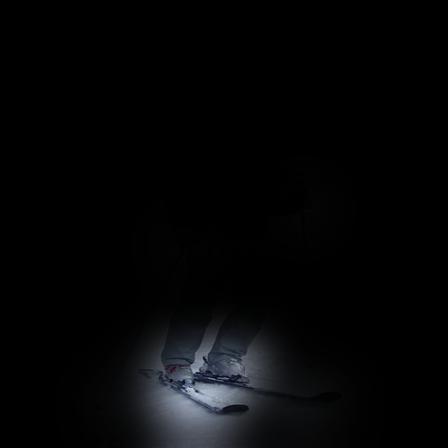}

{\footnotesize \centering {\color[rgb]{1,0.4,0.4}What} {\color[rgb]{1,0,0}color} {\color[rgb]{1,0.5,0.5}are} {\color[rgb]{1,0.5,0.5}the} {\color[rgb]{1,0.9,0.9}skiers} {\color[rgb]{1,0.2,0.2}shoes} \par}
\end{minipage}

{\footnotesize Pred: White, Ans: White\par}
\end{minipage}
}
\end{minipage}

\end{figure*}

\begin{figure*}[hpt!]
\centering

\begin{minipage}{1\textwidth}
\centering

\fcolorbox{green}{white}{
\begin{minipage}{.45\textwidth}
\centering

\begin{minipage}[b]{.45\linewidth}
\includegraphics[width=\linewidth]{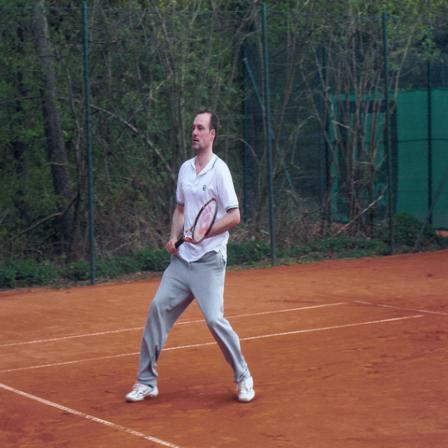}

{\footnotesize \centering Does the man have a beard \par}
\end{minipage}
\begin{minipage}[b]{.45\linewidth}
\includegraphics[width=\linewidth]{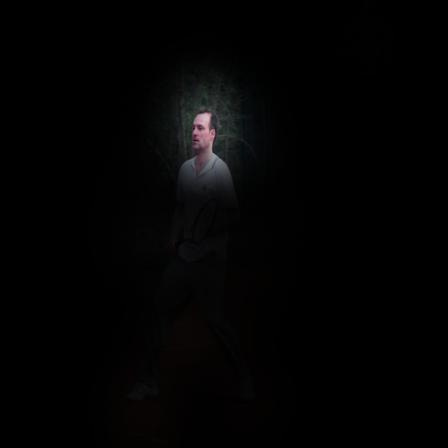}

{\footnotesize \centering {\color[rgb]{1,0.9,0.9}Does} {\color[rgb]{1,0.5,0.5}the} {\color[rgb]{1,0.9,0.9}man} {\color[rgb]{1,0,0}have} {\color[rgb]{1,0.2,0.2}a beard} \par}
\end{minipage}

{\footnotesize Pred: No, Ans: No\par}
\end{minipage}
}
\fcolorbox{green}{white}{
\begin{minipage}{.45\textwidth}
\centering

\begin{minipage}[b]{.45\linewidth}
\includegraphics[width=\linewidth]{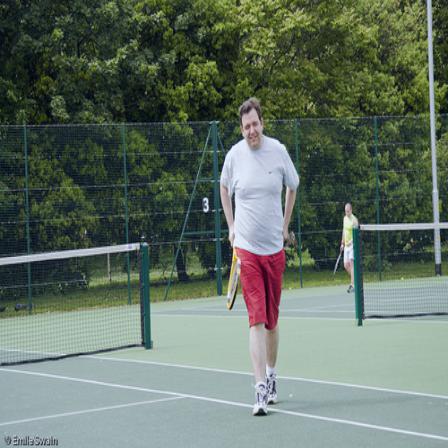}

{\footnotesize \centering Does the man have a beard \par}
\end{minipage}
\begin{minipage}[b]{.45\linewidth}
\includegraphics[width=\linewidth]{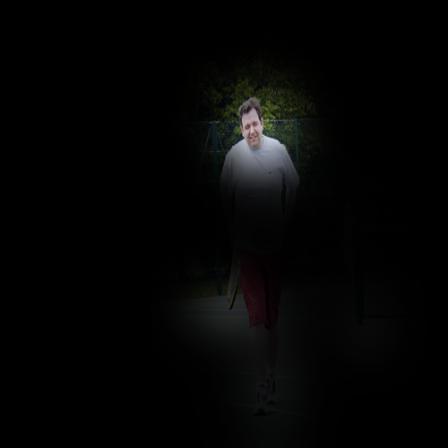}

{\footnotesize \centering {\color[rgb]{1,0.9,0.9}Does} {\color[rgb]{1,0.5,0.5}the} {\color[rgb]{1,0.9,0.9}man} {\color[rgb]{1,0,0}have} {\color[rgb]{1,0.5,0.5}a} {\color[rgb]{1,0.2,0.2}beard} \par}
\end{minipage}

{\footnotesize Pred: No, Ans: No\par}
\end{minipage}
}
\end{minipage}

\smallskip
\begin{minipage}{1\textwidth}
\centering

\fcolorbox{green}{white}{
\begin{minipage}{.45\textwidth}
\centering

\begin{minipage}[b]{.45\linewidth}
\includegraphics[width=\linewidth]{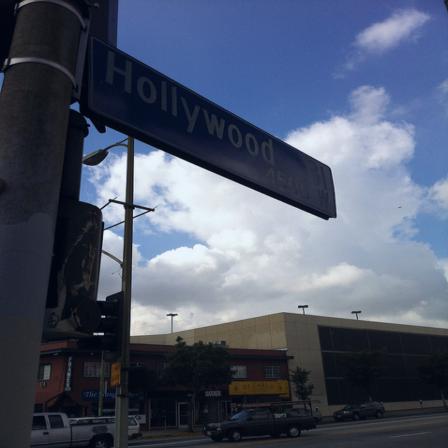}

{\footnotesize \centering Is the sky blue or cloudy \par}
\end{minipage}
\begin{minipage}[b]{.45\linewidth}
\includegraphics[width=\linewidth]{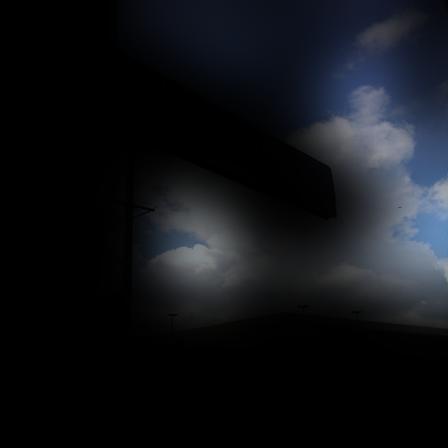}

{\footnotesize \centering {\color[rgb]{1,0.4,0.4}Is} {\color[rgb]{1,0.6,0.6}the sky} {\color[rgb]{1,0,0}blue} {\color[rgb]{1,0.9,0.9}or} {\color[rgb]{1,0.2,0.2}cloudy} \par}
\end{minipage}

{\footnotesize Pred: Cloudy, Ans: Cloudy\par}
\end{minipage}
}
\fcolorbox{green}{white}{
\begin{minipage}{.45\textwidth}
\centering

\begin{minipage}[b]{.45\linewidth}
\includegraphics[width=\linewidth]{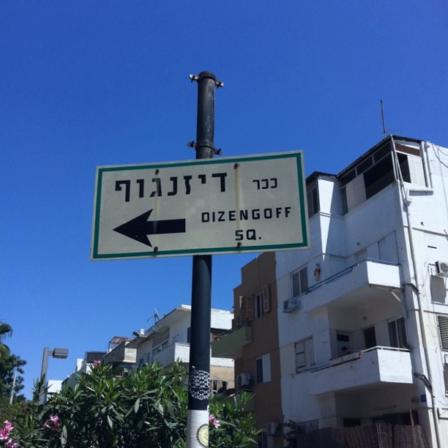}

{\footnotesize \centering Is the sky blue or cloudy  \par}
\end{minipage}
\begin{minipage}[b]{.45\linewidth}
\includegraphics[width=\linewidth]{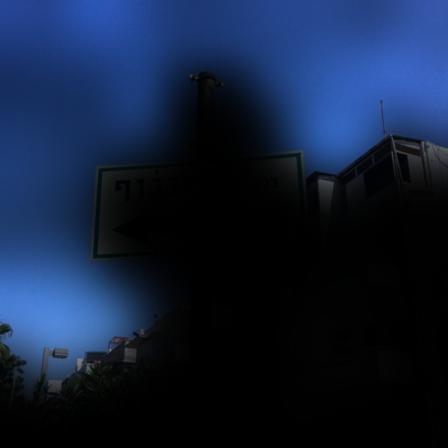}

{\footnotesize \centering {\color[rgb]{1,0.4,0.4}Is} {\color[rgb]{1,0.6,0.6}the sky} {\color[rgb]{1,0.2,0.2}blue} {\color[rgb]{1,0.9,0.9}or} {\color[rgb]{1,0.1,0.1}cloudy} \par}
\end{minipage}

{\footnotesize Pred: Blue, Ans: Blue\par}
\end{minipage}
}
\end{minipage}

\smallskip
\begin{minipage}{1\textwidth}
\centering

\fcolorbox{green}{white}{
\begin{minipage}{.45\textwidth}
\centering

\begin{minipage}[b]{.45\linewidth}
\includegraphics[width=\linewidth]{correct_pairs/pair9a_pic.png}

{\footnotesize \centering How many elephants \par}
\end{minipage}
\begin{minipage}[b]{.45\linewidth}
\includegraphics[width=\linewidth]{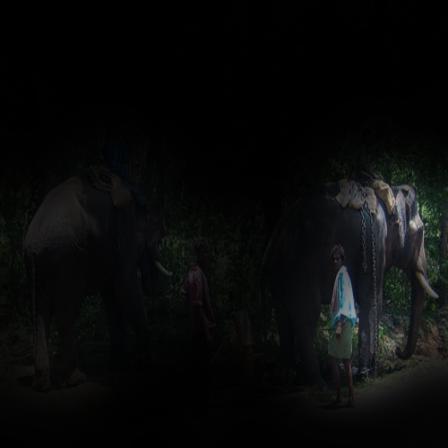}

{\footnotesize \centering {\color[rgb]{1,0.5,0.5}How} {\color[rgb]{1,0,0}many} {\color[rgb]{1,0.9,0.9}elephants} \par}
\end{minipage}

{\footnotesize Pred: 2, Ans: 2\par}
\end{minipage}
}
\fcolorbox{green}{white}{
\begin{minipage}{.45\textwidth}
\centering

\begin{minipage}[b]{.45\linewidth}
\includegraphics[width=\linewidth]{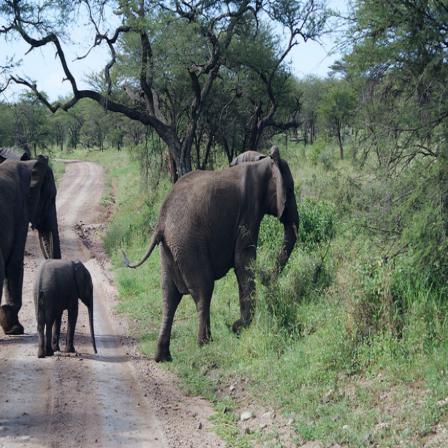}

{\footnotesize \centering How many elephants \par}
\end{minipage}
\begin{minipage}[b]{.45\linewidth}
\includegraphics[width=\linewidth]{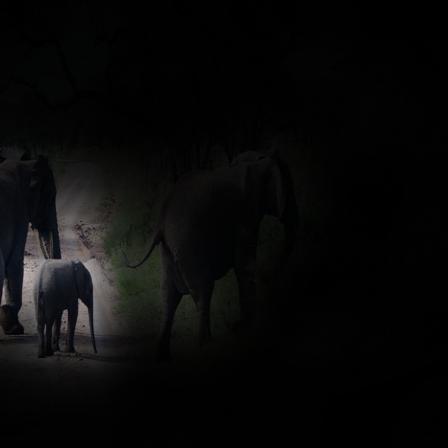}

{\footnotesize \centering {\color[rgb]{1,0.5,0.5}How} {\color[rgb]{1,0,0}many} {\color[rgb]{1,0.9,0.9}elephants} \par}
\end{minipage}

{\footnotesize Pred: 3, Ans: 3\par}
\end{minipage}
}
\end{minipage}

\smallskip
\begin{minipage}{1\textwidth}
\centering

\fcolorbox{green}{white}{
\begin{minipage}{.45\textwidth}
\centering

\begin{minipage}[b]{.45\linewidth}
\includegraphics[width=\linewidth]{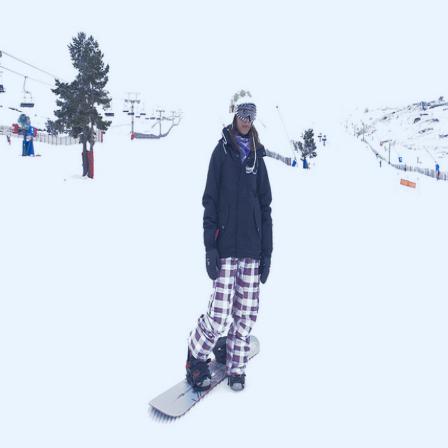}

{\footnotesize \centering What is the women riding \par}
\end{minipage}
\begin{minipage}[b]{.45\linewidth}
\includegraphics[width=\linewidth]{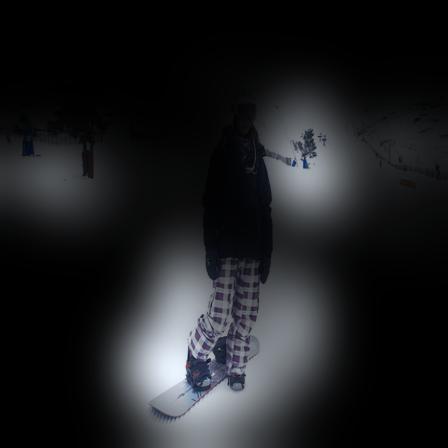}

{\footnotesize \centering {\color[rgb]{1,0,0}What} {\color[rgb]{1,0.3,0.3}is the} {\color[rgb]{1,0.9,0.9}women} {\color[rgb]{1,0.5,0.5}riding} \par}
\end{minipage}

{\footnotesize Pred: Snowboard, Ans: Snowboard\par}
\end{minipage}
}
\fcolorbox{green}{white}{
\begin{minipage}{.45\textwidth}
\centering

\begin{minipage}[b]{.45\linewidth}
\includegraphics[width=\linewidth]{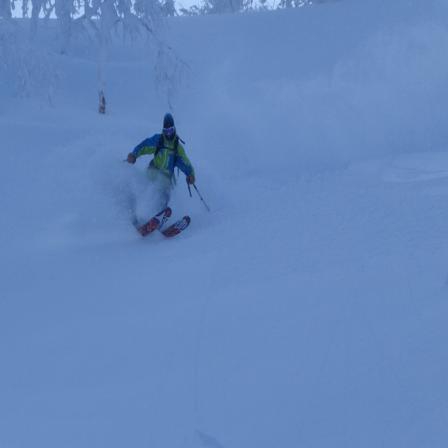}

{\footnotesize \centering What is the women riding \par}
\end{minipage}
\begin{minipage}[b]{.45\linewidth}
\includegraphics[width=\linewidth]{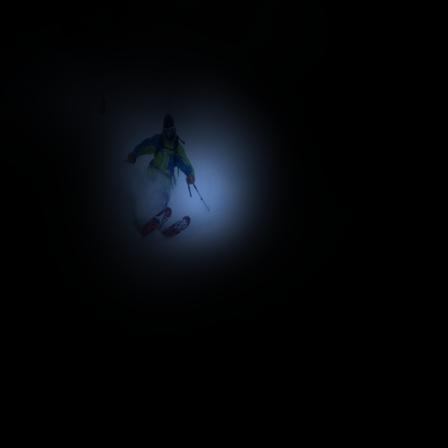}

{\footnotesize \centering {\color[rgb]{1,0,0}What} {\color[rgb]{1,0.4,0.4}is the} {\color[rgb]{1,0.7,0.7}women} {\color[rgb]{1,0,0}riding} \par}
\end{minipage}

{\footnotesize Pred: Skis, Ans: Skis\par}
\end{minipage}
}
\end{minipage}
\end{figure*}

\FloatBarrier
\subsection{Failure Cases}
\label{sec:attn_pred_fail}

According to our analysis, failure cases can be categorized into the following four types:
\begin{itemize}
\setlength{\leftskip}{1.5em}
\item[\em Type-1] Although the DCN is able to locate appropriate image regions and words, it fails to distinguish two different objects or concepts that have similar appearance. This may be attributable to that the extracted image features are not rich enough to distinguish them 
(e.g. {\it mutt} and {\it lab}; and {\it spoon} and {\it fork}).

\item[\em Type-2] Although the DCN is able to locate appropriate image regions and words, it fails to yield correct answers due to the bias of the dataset or missing instances of some objects/concepts in the dataset. For example, there are many samples of an {\it american flag} but no sample of a {\it dragon flag} in the training set. 

\item[\em Type-3] The DCN fails to locate appropriate image regions. This tends to occur when some image regions have similar appearance to the region that the DCN should attend, or the region that it should attend is too small.

\item[\em Type-4] Although the DCN does yield conceptually correct answers, they are not listed in the given set of  answers  in the dataset and thus judged incorrect.
For instance, while the given correct answer is {\it water}, the DCN outputs {\it beach}, which should also be correct, as in one of the examples below.
\end{itemize}

As in the above success cases, each row shows a complementary pair having the same question and different images. In each row, at least either one of the two has an erroneous prediction. The red bounding boxes indicate erroneous answers and the green ones indicate correct answers. The numbers in the failure examples indicate the error types we categorize above. 
\begin{figure*}
\centering

\begin{minipage}{1\textwidth}
\centering

\fcolorbox{red}{white}{
\begin{minipage}{.45\textwidth}
\centering

\begin{minipage}[b]{.45\linewidth}
\includegraphics[width=\linewidth]{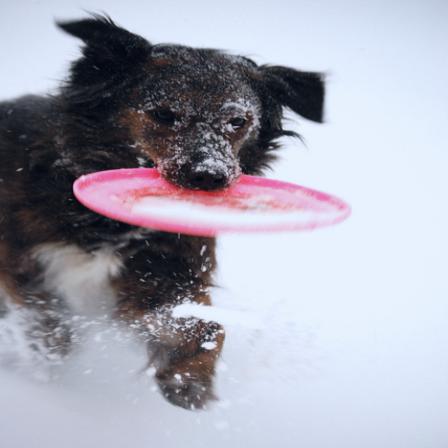}

{\footnotesize \centering What breed of dog is this \par}
\end{minipage}
\begin{minipage}[b]{.45\linewidth}
\includegraphics[width=\linewidth]{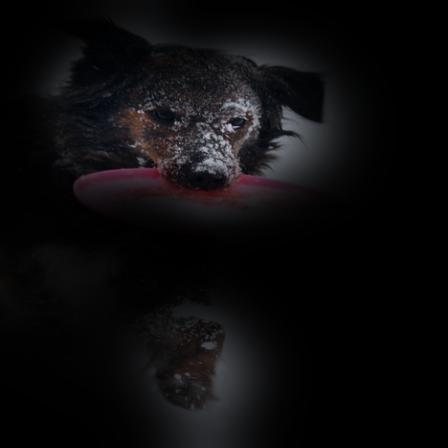}

{\footnotesize \centering {\color[rgb]{1,0.9,0.9}What} {\color[rgb]{1,0.5,0.5}breed} {\color[rgb]{1,0.3,0.3}of} {\color[rgb]{1,0,0}dog} {\color[rgb]{1,0.9,0.9} is this} \par}
\end{minipage}

{\footnotesize Pred: Mutt, Ans: Lab {\em ~~{\color{red}(Error type: 1)}}\par}
\end{minipage}
}
\fcolorbox{green}{white}{
\begin{minipage}{.45\textwidth}
\centering

\begin{minipage}[b]{.45\linewidth}
\includegraphics[width=\linewidth]{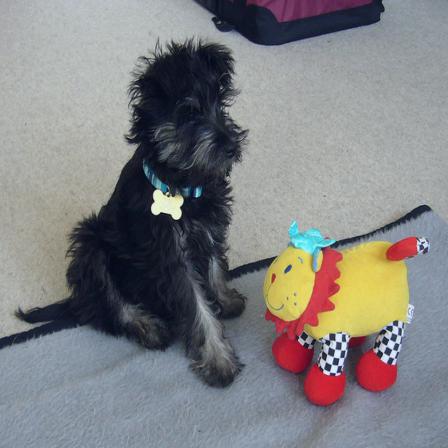}

{\footnotesize \centering What breed of dog is this \par}
\end{minipage}
\begin{minipage}[b]{.45\linewidth}
\includegraphics[width=\linewidth]{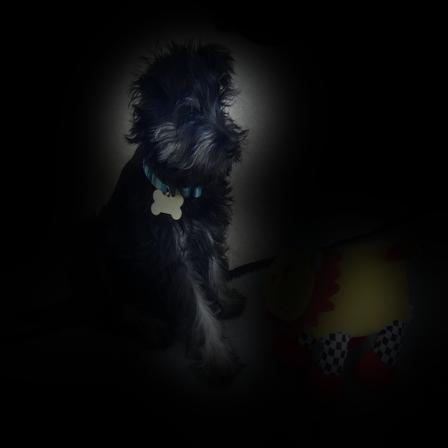}

{\footnotesize \centering {\color[rgb]{1,0.9,0.9}What} {\color[rgb]{1,0.6,0.6}breed} {\color[rgb]{1,0.4,0.4}of} {\color[rgb]{1,0,0}dog} {\color[rgb]{1,0.9,0.9} is this} \par}
\end{minipage}

{\footnotesize Pred: Terrier, Ans: Terrier\par} 
\end{minipage}
}
\end{minipage}

\smallskip
\begin{minipage}{1\textwidth}
\centering

\fcolorbox{green}{white}{
\begin{minipage}{.45\textwidth}
\centering

\begin{minipage}[b]{.45\linewidth}
\includegraphics[width=\linewidth]{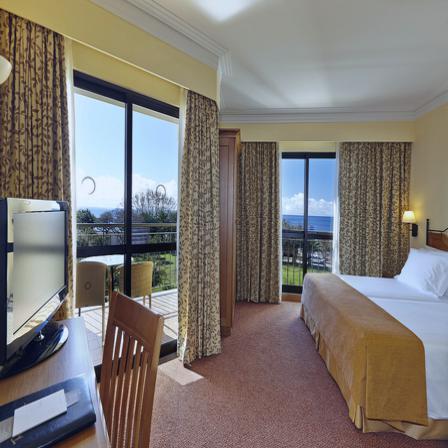}

{\footnotesize \centering What room is this \par}
\end{minipage}
\begin{minipage}[b]{.45\linewidth}
\includegraphics[width=\linewidth]{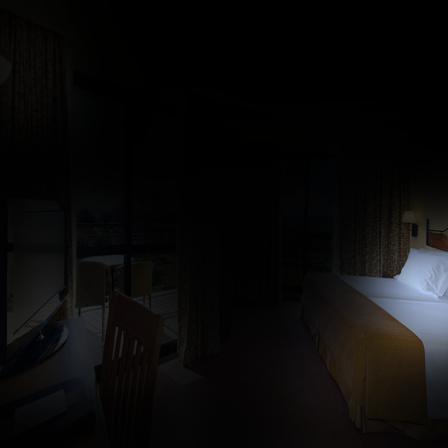}

{\footnotesize \centering {\color[rgb]{1,0.6,0.6}What} {\color[rgb]{1,0,0}room} {\color[rgb]{1,0.9,0.9}is} {\color[rgb]{1,0.7,0.7}this} \par}
\end{minipage}

{\footnotesize Pred: Bedroom, Ans: Bedroom\par}
\end{minipage}
}
\fcolorbox{red}{white}{
\begin{minipage}{.45\textwidth}
\centering

\begin{minipage}[b]{.45\linewidth}
\includegraphics[width=\linewidth]{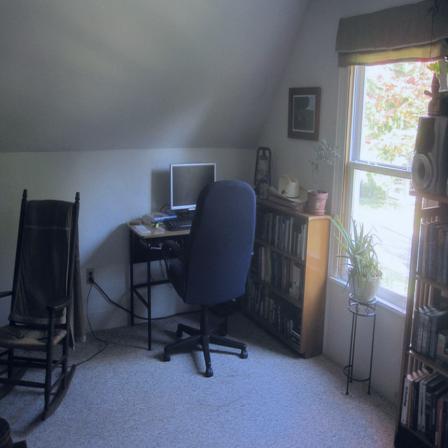}

{\footnotesize \centering What room is this \par}
\end{minipage}
\begin{minipage}[b]{.45\linewidth}
\includegraphics[width=\linewidth]{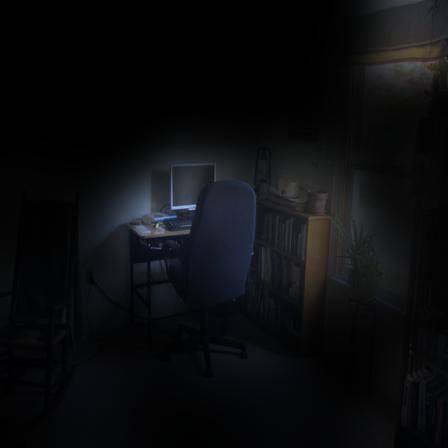}

{\footnotesize \centering {\color[rgb]{1,0.6,0.6}What} {\color[rgb]{1,0,0}room} {\color[rgb]{1,0.9,0.9}is} {\color[rgb]{1,0.9,0.9}this} \par}
\end{minipage}

{\footnotesize Pred: Living room, Ans: Office ~~{\color{red} {\em (Error type: 1)}}\par}
\end{minipage}
}
\end{minipage}

\smallskip
\begin{minipage}{1\textwidth}
\centering

\fcolorbox{green}{white}{
\begin{minipage}{.45\textwidth}
\centering

\begin{minipage}[b]{.45\linewidth}
\includegraphics[width=\linewidth]{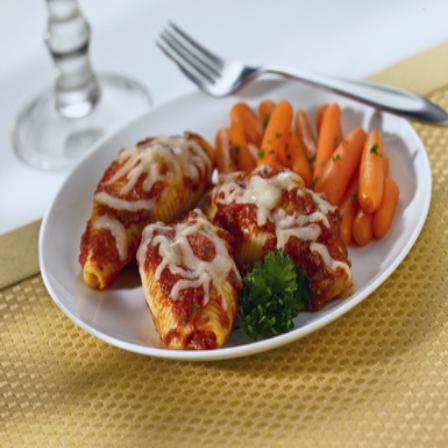}

{\footnotesize \centering What is the name of the utensil \par}
\end{minipage}
\begin{minipage}[b]{.45\linewidth}
\includegraphics[width=\linewidth]{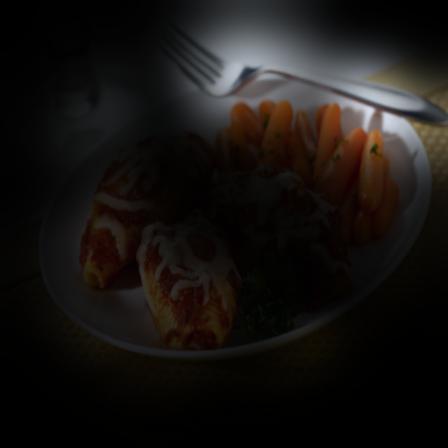}

{\footnotesize \centering {\color[rgb]{1,0.9,0.9}What} {\color[rgb]{1,0.5,0.5}is} {\color[rgb]{1,0.9,0.9}the} {\color[rgb]{1,0.5,0.5}name} {\color[rgb]{1,0.9,0.9}of} {\color[rgb]{1,0.5,0.5}the} {\color[rgb]{1,0,0}utensil} \par}
\end{minipage}

{\footnotesize Pred: Fork, Ans: Fork\par}
\end{minipage}
}
\fcolorbox{red}{white}{
\begin{minipage}{.45\textwidth}
\centering

\begin{minipage}[b]{.45\linewidth}
\includegraphics[width=\linewidth]{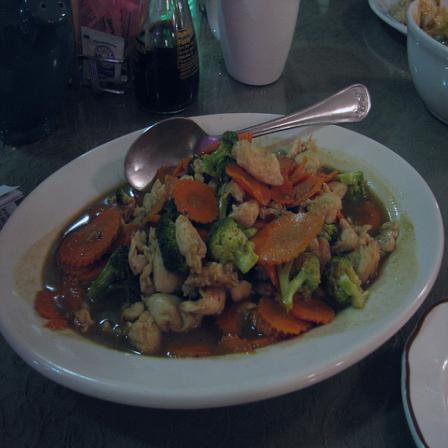}

{\footnotesize \centering What is the name of the utensil \par}
\end{minipage}
\begin{minipage}[b]{.45\linewidth}
\includegraphics[width=\linewidth]{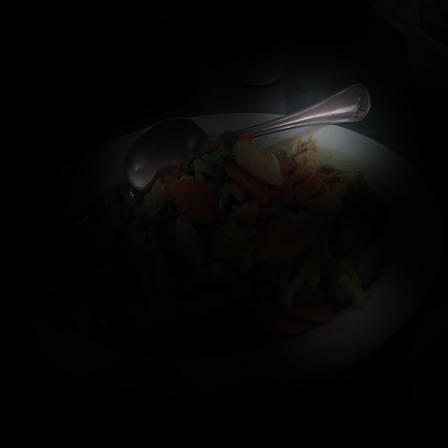}

{\footnotesize \centering {\color[rgb]{1,0.9,0.9}What} {\color[rgb]{1,0.5,0.5}is} {\color[rgb]{1,0.9,0.9}the} {\color[rgb]{1,0.5,0.5}name} {\color[rgb]{1,0.9,0.9}of} {\color[rgb]{1,0.5,0.5}the} {\color[rgb]{1,0,0}utensil} \par}
\end{minipage}

{\footnotesize Pred: Fork, Ans: Spoon ~~{\color{red} {\em (Error type: 1)}}\par}
\end{minipage}
}
\end{minipage}

\smallskip
\begin{minipage}{1\textwidth}
\centering

\fcolorbox{red}{white}{
\begin{minipage}{.45\textwidth}
\centering

\begin{minipage}[b]{.45\linewidth}
\includegraphics[width=\linewidth]{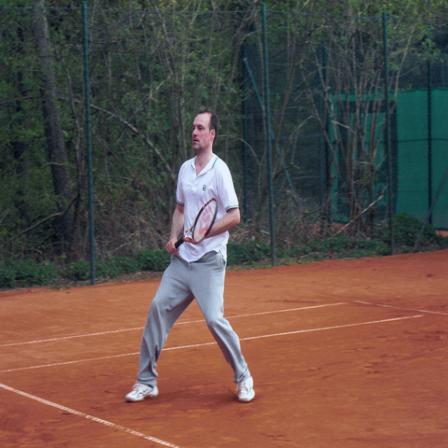}

{\footnotesize \centering How tall is he \par}
\end{minipage}
\begin{minipage}[b]{.45\linewidth}
\includegraphics[width=\linewidth]{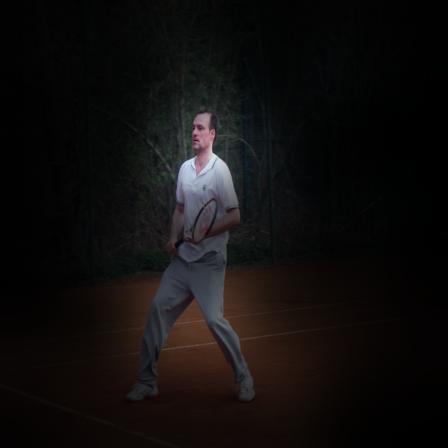}

{\footnotesize \centering {\color[rgb]{1,0.3,0.3}How} {\color[rgb]{1,0,0}tall} {\color[rgb]{1,0.6,0.6}is} {\color[rgb]{1,0.9,0.9}he} \par}
\end{minipage}

{\footnotesize Pred: 5 feet, Ans: Tall ~~{\color{red} {\em (Error type: 1)}}\par}
\end{minipage}
}
\fcolorbox{red}{white}{
\begin{minipage}{.45\textwidth}
\centering

\begin{minipage}[b]{.45\linewidth}
\includegraphics[width=\linewidth]{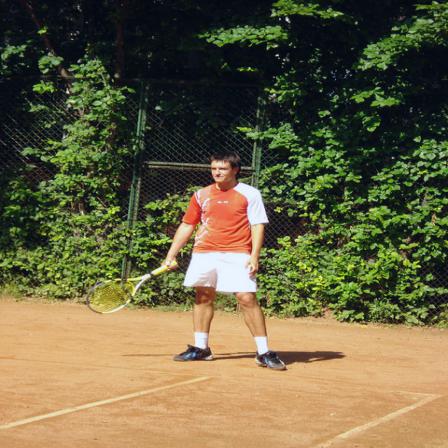}

{\footnotesize \centering How tall is he \par}
\end{minipage}
\begin{minipage}[b]{.45\linewidth}
\includegraphics[width=\linewidth]{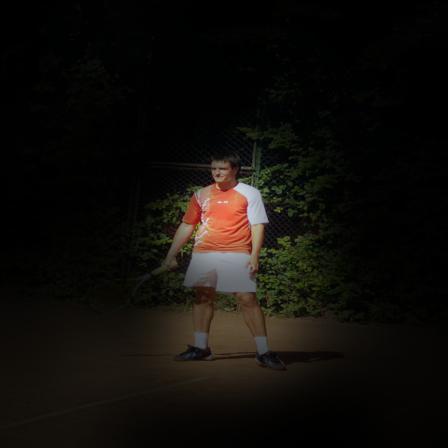}

{\footnotesize \centering {\color[rgb]{1,0.3,0.3}How} {\color[rgb]{1,0,0}tall} {\color[rgb]{1,0.6,0.6}is} {\color[rgb]{1,0.9,0.9}he} \par}
\end{minipage}

{\footnotesize Pred: 5 feet, Ans: 6 feet ~~{\color{red} {\em (Error type: 2)}}\par}
\end{minipage}
}
\end{minipage}

\end{figure*}

\begin{figure*}[ht]
\centering

\begin{minipage}{1\textwidth}
\centering

\smallskip
\fcolorbox{red}{white}{
\begin{minipage}{.45\textwidth}
\centering

\begin{minipage}[b]{.45\linewidth}
\includegraphics[width=\linewidth]{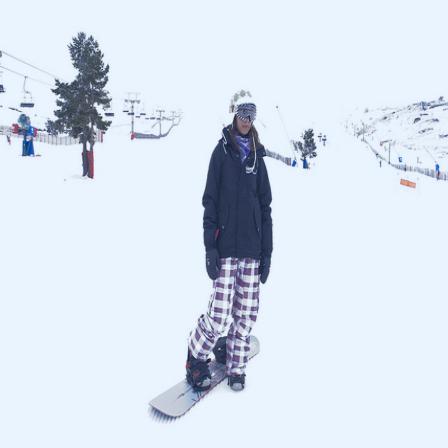}

{\footnotesize \centering What is the color of pants the woman is wearing \par}
\end{minipage}
\begin{minipage}[b]{.45\linewidth}
\includegraphics[width=\linewidth]{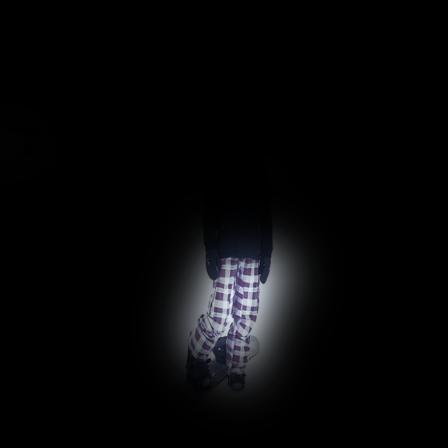}

{\footnotesize \centering {\color[rgb]{1,0.9,0.9}What} {\color[rgb]{1,0.6,0.6}is} {\color[rgb]{1,0.9,0.9}the} {\color[rgb]{1,0,0}color} {\color[rgb]{1,0.9,0.9}of} {\color[rgb]{1,0.2,0.2}pants} {\color[rgb]{1,0.9,0.9}the woman is wearing} \par}
\end{minipage}

{\footnotesize Pred: Plaid, Ans: Red and White ~~{\color{red} {\em (Error type: 4)}}\par}
\end{minipage}
}
\fcolorbox{red}{white}{
\begin{minipage}{.45\textwidth}
\centering

\begin{minipage}[b]{.45\linewidth}
\includegraphics[width=\linewidth]{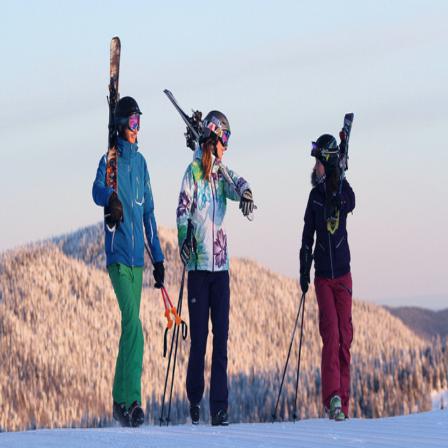}

{\footnotesize \centering What is the color of pants the woman is wearing \par}
\end{minipage}
\begin{minipage}[b]{.45\linewidth}
\includegraphics[width=\linewidth]{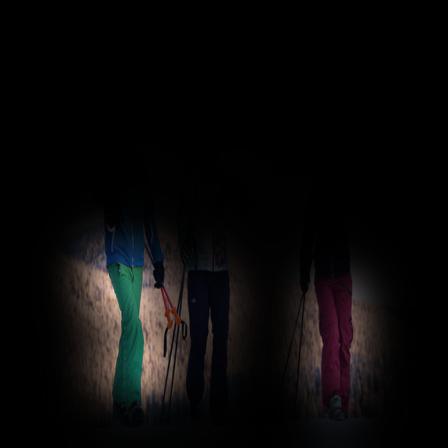}

{\footnotesize \centering {\color[rgb]{1,0.9,0.9}What} {\color[rgb]{1,0.6,0.6}is} {\color[rgb]{1,0.9,0.9}the} {\color[rgb]{1,0,0}color} {\color[rgb]{1,0.9,0.9}of} {\color[rgb]{1,0.2,0.2}pants} {\color[rgb]{1,0.9,0.9}the woman is wearing} \par}
\end{minipage}

{\footnotesize Pred: Green, Ans: Black ~~{\color{red} {\em (Error type: 4)}}\par}
\end{minipage}
}
\end{minipage}

\smallskip
\begin{minipage}{1\textwidth}
\centering

\fcolorbox{red}{white}{
\begin{minipage}{.45\textwidth}
\centering

\begin{minipage}[b]{.45\linewidth}
\includegraphics[width=\linewidth]{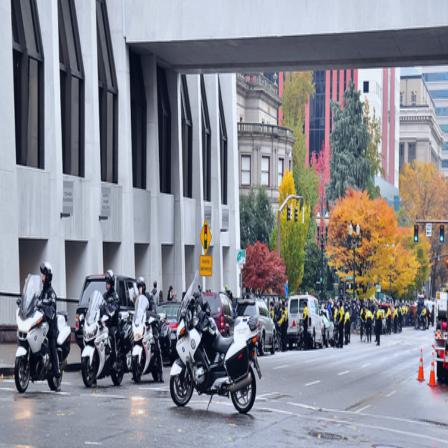}

{\footnotesize \centering What color is lit up on the street lights \par}
\end{minipage}
\begin{minipage}[b]{.45\linewidth}
\includegraphics[width=\linewidth]{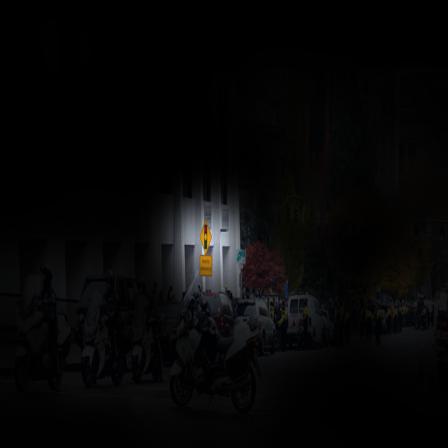}

{\footnotesize \centering {\color[rgb]{1,0.5,0.5}What} {\color[rgb]{1,0,0}color} {\color[rgb]{1,0.7,0.7}is} {\color[rgb]{1,0.5,0.5}lit up} {\color[rgb]{1,0.9,0.9}on the street} {\color[rgb]{1,0.9,0.9}lights} \par}
\end{minipage}

{\footnotesize Pred: Yellow, Ans: Green ~~{\color{red} {\em (Error type: 3)}}\par}
\end{minipage}
}
\fcolorbox{red}{white}{
\begin{minipage}{.45\textwidth}
\centering

\begin{minipage}[b]{.45\linewidth}
\includegraphics[width=\linewidth]{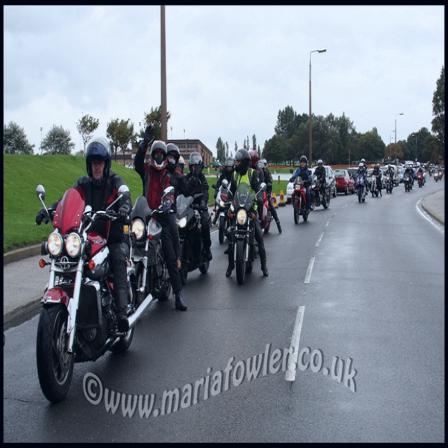}

{\footnotesize \centering What color is lit up on the street lights \par}
\end{minipage}
\begin{minipage}[b]{.45\linewidth}
\includegraphics[width=\linewidth]{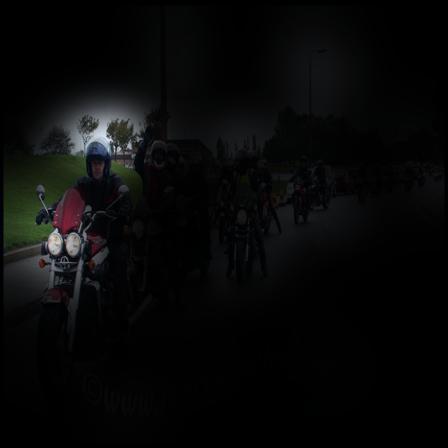}

{\footnotesize \centering {\color[rgb]{1,0.5,0.5}What} {\color[rgb]{1,0,0}color} {\color[rgb]{1,0.7,0.7}is} {\color[rgb]{1,0.5,0.5}lit up} {\color[rgb]{1,0.9,0.9}on the street} {\color[rgb]{1,0.5,0.5}lights} \par}
\end{minipage}

{\footnotesize Pred: White, Ans: None ~~{\color{red} {\em (Error type: 1)}}\par}
\end{minipage}
}
\end{minipage}

\smallskip
\begin{minipage}{1\textwidth}
\centering

\fcolorbox{red}{white}{
\begin{minipage}{.45\textwidth}
\centering

\begin{minipage}[b]{.45\linewidth}
\includegraphics[width=\linewidth]{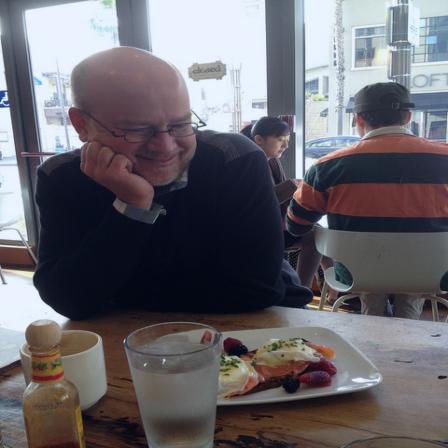}

{\footnotesize \centering Where is the fruit \par}
\end{minipage}
\begin{minipage}[b]{.45\linewidth}
\includegraphics[width=\linewidth]{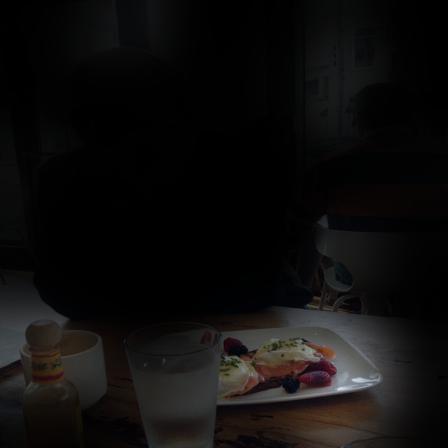}

{\footnotesize \centering {\color[rgb]{1,0,0}Where} {\color[rgb]{1,0.9,0.9}is} {\color[rgb]{1,0.6,0.6}the} {\color[rgb]{1,0.9,0.9}fruit} \par}
\end{minipage}

{\footnotesize Pred: Table, Ans: Plate ~~{\color{red} {\em (Error type: 4)}}\par}
\end{minipage}
}
\fcolorbox{green}{white}{
\begin{minipage}{.45\textwidth}
\centering

\begin{minipage}[b]{.45\linewidth}
\includegraphics[width=\linewidth]{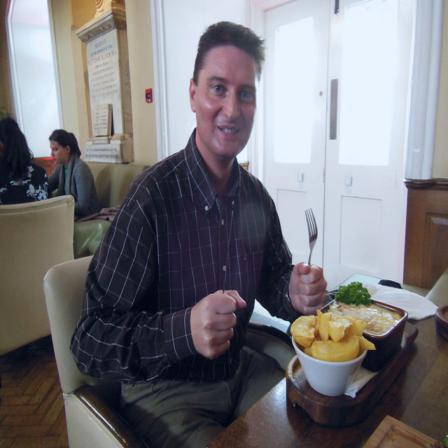}

{\footnotesize \centering Where is the fruit \par}
\end{minipage}
\begin{minipage}[b]{.45\linewidth}
\includegraphics[width=\linewidth]{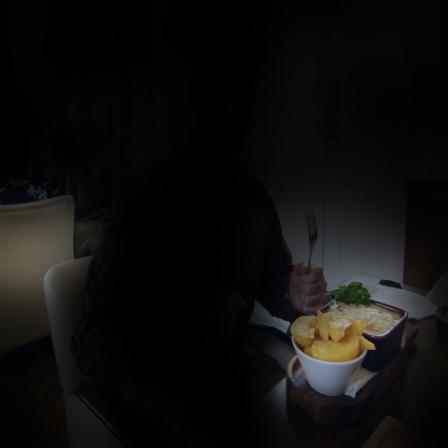}

{\footnotesize \centering {\color[rgb]{1,0,0}Where} {\color[rgb]{1,0.9,0.9}is} {\color[rgb]{1,0.6,0.6}the} {\color[rgb]{1,0.9,0.9}fruit} \par}
\end{minipage}

{\footnotesize Pred: Bowl, Ans: Bowl\par}
\end{minipage}
}
\end{minipage}

\smallskip
\begin{minipage}{1\textwidth}
\centering

\fcolorbox{red}{white}{
\begin{minipage}{.45\textwidth}
\centering

\begin{minipage}[b]{.45\linewidth}
\includegraphics[width=\linewidth]{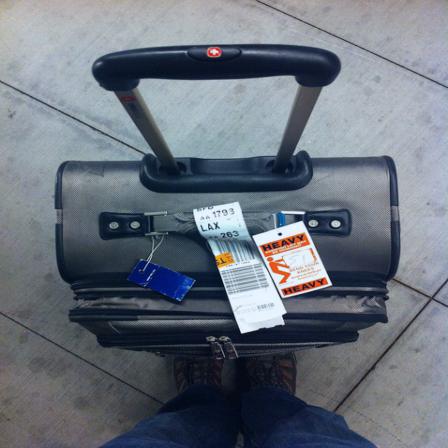}

{\footnotesize \centering How many tags are on the suitcase \par}
\end{minipage}
\begin{minipage}[b]{.45\linewidth}
\includegraphics[width=\linewidth]{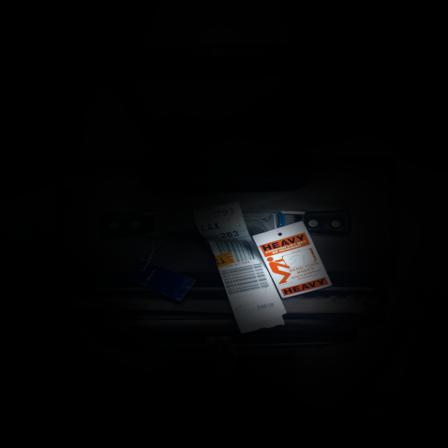}

{\footnotesize \centering {\color[rgb]{1,0.3,0.3}How} {\color[rgb]{1,0,0}many} {\color[rgb]{1,0.5,0.5}tags} {\color[rgb]{1,0.9,0.9}are} {\color[rgb]{1,0.9,0.9}on the suitcase} \par}
\end{minipage}

{\footnotesize Pred: 4, Ans: 3 ~~{\color{red} {\em (Error type: 1)}}\par}
\end{minipage}
}
\fcolorbox{green}{white}{
\begin{minipage}{.45\textwidth}
\centering

\begin{minipage}[b]{.45\linewidth}
\includegraphics[width=\linewidth]{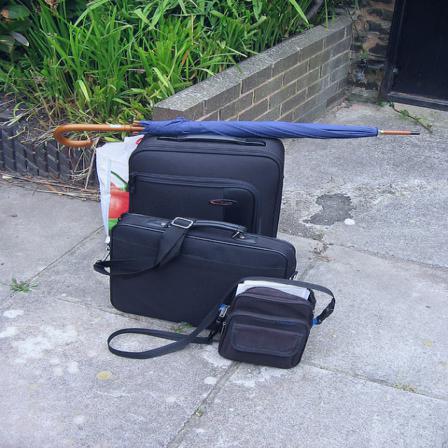}

{\footnotesize \centering How many tags are on the suitcase \par}
\end{minipage}
\begin{minipage}[b]{.45\linewidth}
\includegraphics[width=\linewidth]{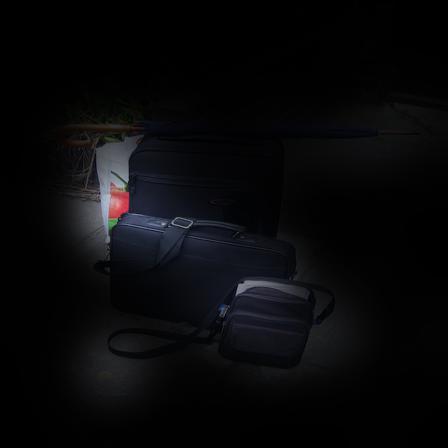}

{\footnotesize \centering {\color[rgb]{1,0.4,0.4}How} {\color[rgb]{1,0,0}many} {\color[rgb]{1,0.4,0.4}tags} {\color[rgb]{1,0.5,0.5}are} {\color[rgb]{1,0.9,0.9}on the suitcase} \par}
\end{minipage}

{\footnotesize Pred: 0, Ans: 0\par}
\end{minipage}
}
\end{minipage}

\end{figure*}

\begin{figure*}[ht]
\centering
\begin{minipage}{1\textwidth}
\centering

\fcolorbox{green}{white}{
\begin{minipage}{.45\textwidth}
\centering

\begin{minipage}[b]{.45\linewidth}
\includegraphics[width=\linewidth]{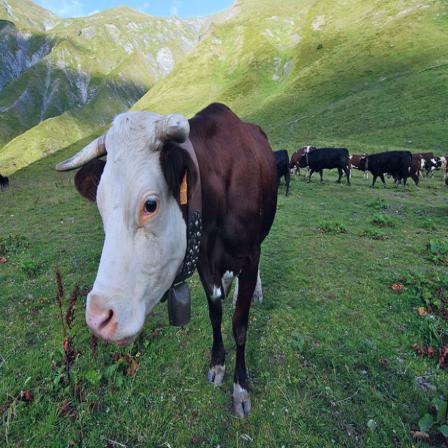}

{\footnotesize \centering What landforms are behind the cows \par}
\end{minipage}
\begin{minipage}[b]{.45\linewidth}
\includegraphics[width=\linewidth]{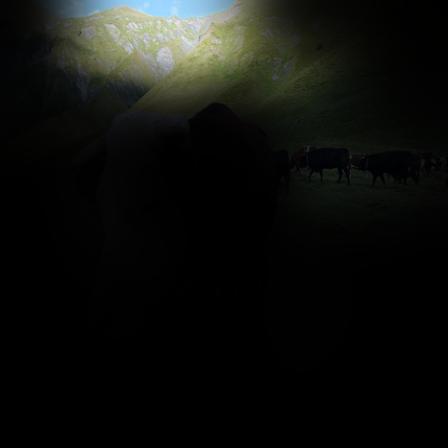}

{\footnotesize \centering {\color[rgb]{1,0.9,0.9}What} {\color[rgb]{1,0,0}landforms} {\color[rgb]{1,0.5,0.5}are} {\color[rgb]{1,0.2,0.2}behind} {\color[rgb]{1,0.9,0.9}the cows} \par}
\end{minipage}

{\footnotesize Pred: Mountains, Ans: Mountains\par}
\end{minipage}
}
\fcolorbox{red}{white}{
\begin{minipage}{.45\textwidth}
\centering

\begin{minipage}[b]{.45\linewidth}
\includegraphics[width=\linewidth]{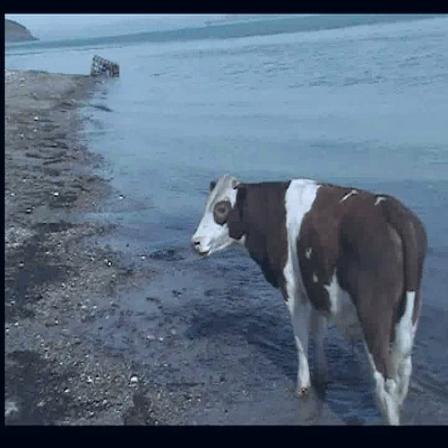}

{\footnotesize \centering What landforms are behind the cows \par}
\end{minipage}
\begin{minipage}[b]{.45\linewidth}
\includegraphics[width=\linewidth]{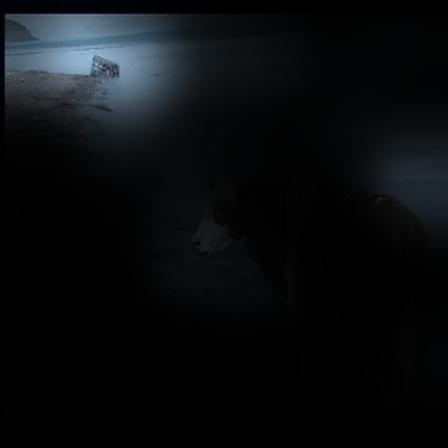}

{\footnotesize \centering {\color[rgb]{1,0.5,0.5}What} {\color[rgb]{1,0,0}landforms} {\color[rgb]{1,0.5,0.5}are} {\color[rgb]{1,0.2,0.2}behind} {\color[rgb]{1,0.9,0.9}the cows} \par}
\end{minipage}

{\footnotesize Pred: Beach, Ans: Water ~~{\color{red} {\em (Error type: 4)}}\par}
\end{minipage}
}
\end{minipage}

\smallskip
\begin{minipage}{1\textwidth}
\centering

\fcolorbox{green}{white}{
\begin{minipage}{.45\textwidth}
\centering

\begin{minipage}[b]{.45\linewidth}
\includegraphics[width=\linewidth]{wrong_pairs/pair7a_pic.png}

{\footnotesize \centering What flag is that \par}
\end{minipage}
\begin{minipage}[b]{.45\linewidth}
\includegraphics[width=\linewidth]{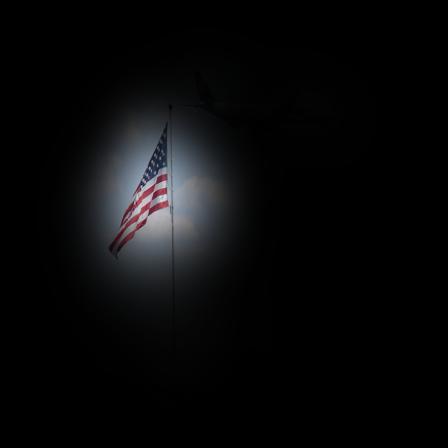}

{\footnotesize \centering {\color[rgb]{1,0.9,0.9}What} {\color[rgb]{1,0,0}flag} {\color[rgb]{1,0.9,0.9}is that} \par}
\end{minipage}

{\footnotesize Pred: American, Ans: American\par}
\end{minipage}
}
\fcolorbox{red}{white}{
\begin{minipage}{.45\textwidth}
\centering

\begin{minipage}[b]{.45\linewidth}
\includegraphics[width=\linewidth]{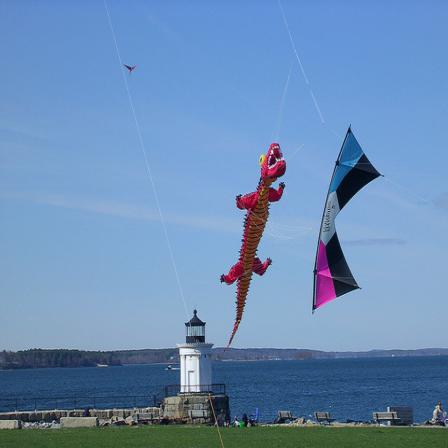}

{\footnotesize \centering What flag is that \par}
\end{minipage}
\begin{minipage}[b]{.45\linewidth}
\includegraphics[width=\linewidth]{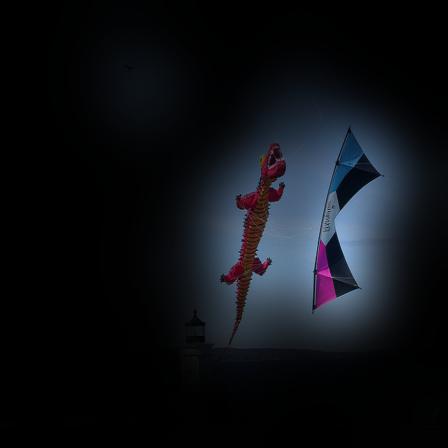}

{\footnotesize \centering {\color[rgb]{1,0.9,0.9}What} {\color[rgb]{1,0,0}flag} {\color[rgb]{1,0.9,0.9}is that} \par}
\end{minipage}

{\footnotesize Pred: American, Ans: Dragon ~~{\color{red} {\em (Error type: 2)}}\par}
\end{minipage}
}
\end{minipage}

\smallskip
\begin{minipage}{1\textwidth}
\centering

\fcolorbox{green}{white}{
\begin{minipage}{.45\textwidth}
\centering

\begin{minipage}[b]{.45\linewidth}
\includegraphics[width=\linewidth]{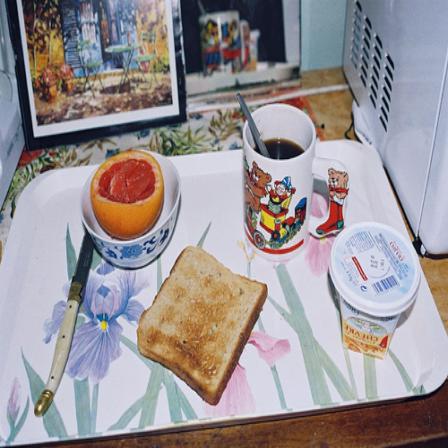}

{\footnotesize \centering What is in the mug \par}
\end{minipage}
\begin{minipage}[b]{.45\linewidth}
\includegraphics[width=\linewidth]{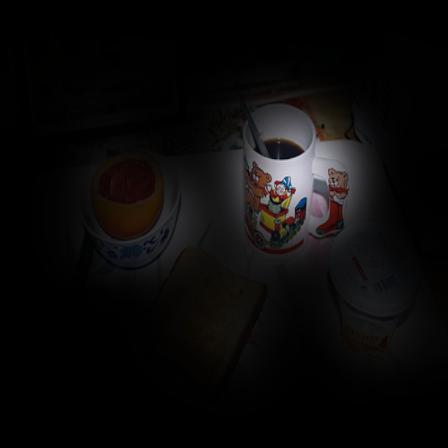}

{\footnotesize {\color[rgb]{1,0.4,0.4}What} {\color[rgb]{1,0.2,0.2}is} {\color[rgb]{1,0.9,0.9}in} {\color[rgb]{1,0.9,0.9}the} {\color[rgb]{1,0,0}mug} \par}
\end{minipage}

{\footnotesize Pred: Coffee, Ans: Coffee\par}
\end{minipage}
}
\fcolorbox{red}{white}{
\begin{minipage}{.45\textwidth}
\centering

\begin{minipage}[b]{.45\linewidth}
\includegraphics[width=\linewidth]{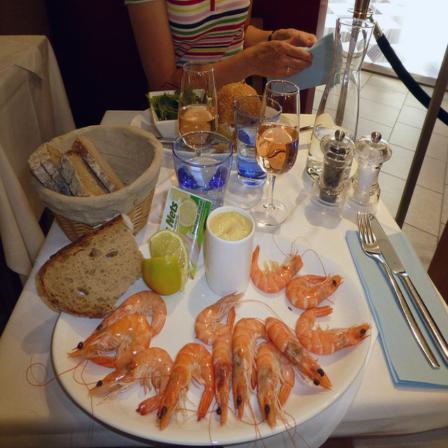}

{\footnotesize \centering What is in the mug \par}
\end{minipage}
\begin{minipage}[b]{.45\linewidth}
\includegraphics[width=\linewidth]{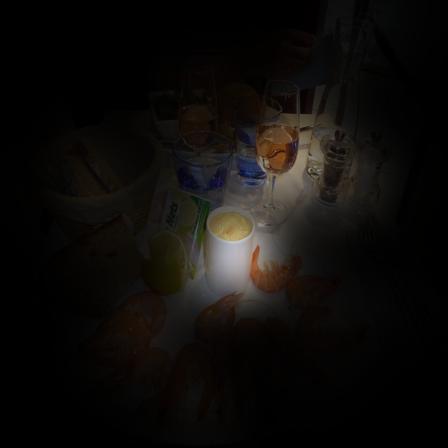}

{\footnotesize {\color[rgb]{1,0.4,0.4}What} {\color[rgb]{1,0.3,0.3}is} {\color[rgb]{1,0.9,0.9}in} {\color[rgb]{1,0.9,0.9}the} {\color[rgb]{1,0,0}mug} \par}
\end{minipage}

{\footnotesize Pred: Wine, Ans: Butter ~~{\color{red} {\em (Error type: 1)}}\par}
\end{minipage}
}
\end{minipage}

\smallskip
\begin{minipage}{1\textwidth}
\centering

\fcolorbox{red}{white}{
\begin{minipage}{.45\textwidth}
\centering

\begin{minipage}[b]{.45\linewidth}
\includegraphics[width=\linewidth]{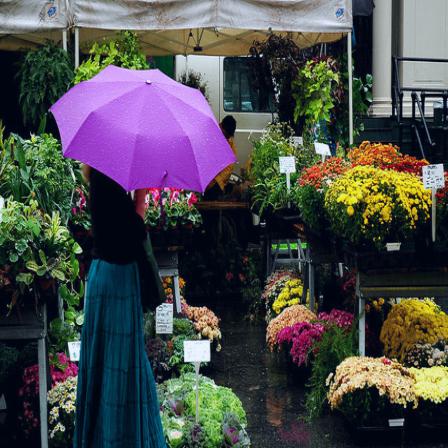}

{\footnotesize \centering Where is this woman at \par}
\end{minipage}
\begin{minipage}[b]{.45\linewidth}
\includegraphics[width=\linewidth]{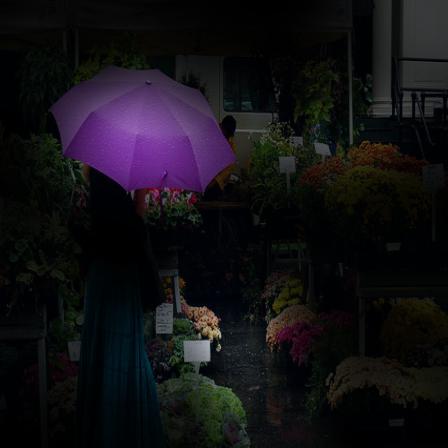}

{\footnotesize \centering {\color[rgb]{1,0,0}Where} {\color[rgb]{1,0.3,0.3}is} {\color[rgb]{1,0.6,0.6}this} {\color[rgb]{1,0.9,0.9}woman} {\color[rgb]{1,0.6,0.6}at} \par}
\end{minipage}

{\footnotesize Pred: Outside, Ans: Farmers market  ~~{\color{red} {\em (Error type: 4)}}\par}
\end{minipage}
}
\fcolorbox{green}{white}{
\begin{minipage}{.45\textwidth}
\centering

\begin{minipage}[b]{.45\linewidth}
\includegraphics[width=\linewidth]{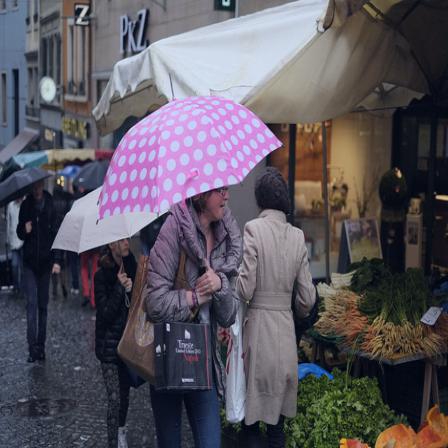}

{\footnotesize \centering Where is this woman at \par}
\end{minipage}
\begin{minipage}[b]{.45\linewidth}
\includegraphics[width=\linewidth]{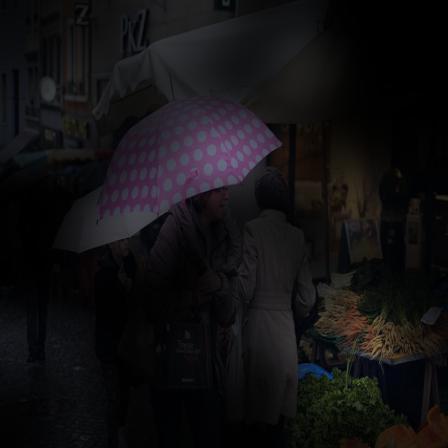}

{\footnotesize \centering {\color[rgb]{1,0.2,0.2}Where} {\color[rgb]{1,0.3,0.3}is} {\color[rgb]{1,0.6,0.6}this} {\color[rgb]{1,0.9,0.9}woman} {\color[rgb]{1,0.4,0.4}at} \par}
\end{minipage}

{\footnotesize Pred: Market, Ans: Market\par}
\end{minipage}
}
\end{minipage}
\end{figure*}

\section{
Layer Attention in the Image Feature Extraction Step
}
\label{sec:result_extrac}

As explained in the main paper (Sec.3.1), our DCN extracts visual features from an input image using a pre-trained ResNet at the initial step. The features are obtained by computing the weighted sum of the activations (i.e., outputs) of the four convolutional layers of the ResNet, where the attention weights generated conditioned on the input question are used. We examine here how this attention mechanism works for different types of questions. Specifically, utilizing the fifty five question types provided in the VQA-2.0, we compute the mean and standard deviation of the four attention weights for the questions belonging to each question type. We used all the questions in the validation set and our DCN trained only on train set for this computation.

Figure \ref{att4layer} shows the results. The bars in four colors represent the means of the four layer weights for each question type, and the thin black bars attached to the color bars represent their standard deviations. The fifty five question types are ordered by their similarity 
in the horizontal axis. From the plot, we can make the following observations:
\begin{itemize}
    \item Layer 1 (the lowest one) has a certain level of weights only for {\em Yes/No} questions (shown on about the left half of the plots)  and no weight for other types of questions (on the right half);
   \item Layer 2 has  a small weight only for {\em Yes/No} questions and no weight for other types of questions;    \item Layer 3 tends to have large weights for questions about colors (e.g., {\em ``what color''}) and questions about presence of a given object(s) (e.g., {\em ``are there''} and {\em ``how many''});
    \item Layer 4 (the highest one) has the largest attention weights in most of the question types, indicating its importance in answering them.
    \item Specific questions, such as {\em ``what color ''} and {\em ``what sport is''}, tend to have smaller standard deviations than nonspecific questions, such as {\em ``is the woman''} and {\em ``do you''}. 
\end{itemize}

\begin{figure}[t]
\centering
\includegraphics[width=1\linewidth]{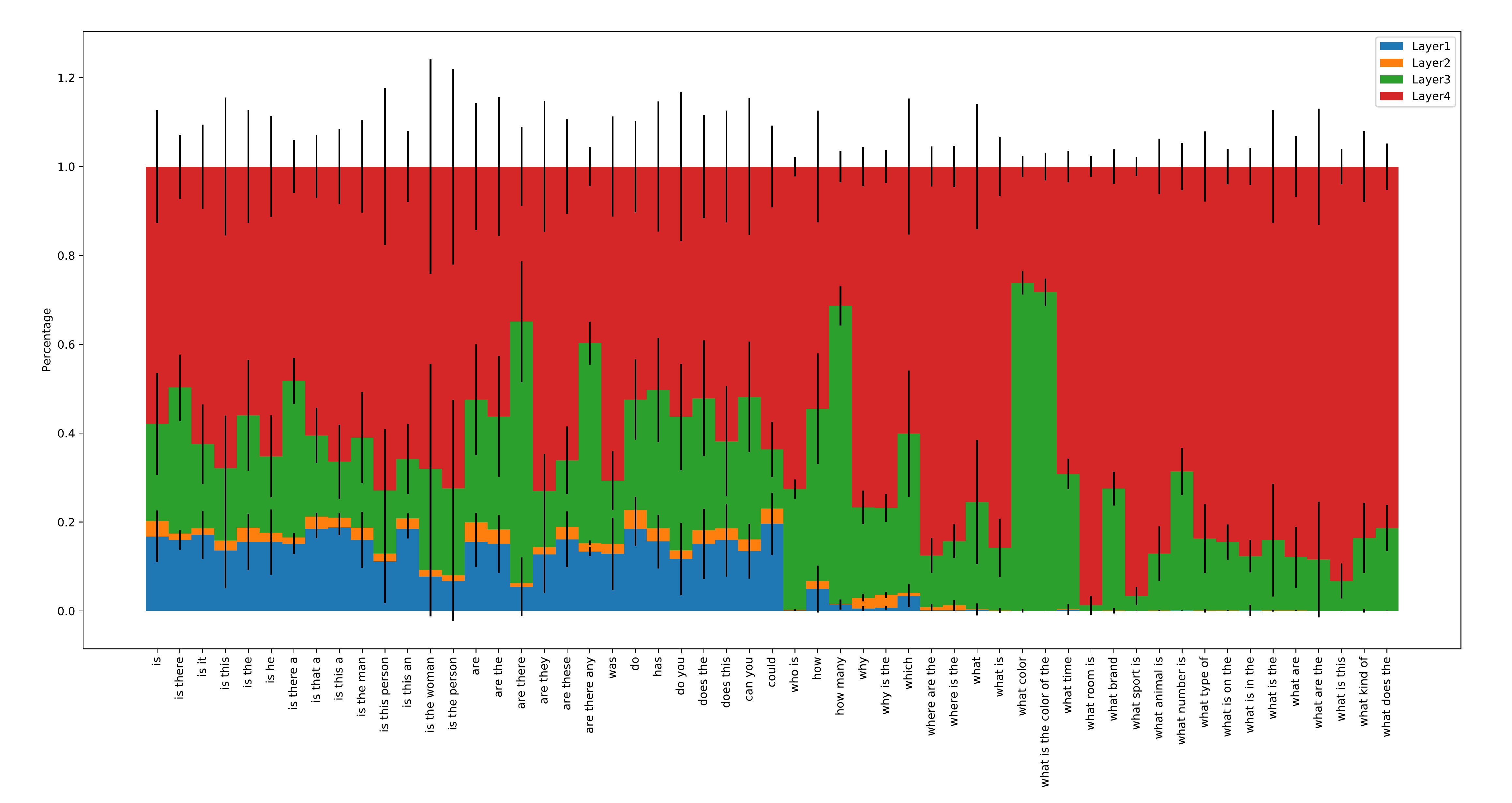}
\setcounter{figure}{5}
\caption{Statistics (means and standard deviations) of the attention weights on the four convolutional layers generated in the image feature extraction step for different types of questions. }
\label{att4layer}
\end{figure}

{\small
\bibliographystyle{ieee}
\bibliography{egbib}
}

\end{document}